\documentclass[10pt,journal,compsoc]{IEEEtran}
\usepackage{times}
\usepackage{epsfig}
\usepackage{graphicx}
\usepackage{multirow}
\usepackage{subcaption}
\usepackage{amsmath}
\usepackage{amssymb}
\usepackage{flushend}
\usepackage{changepage}
\usepackage{makecell}
\usepackage[margin=0.8245in]{geometry}

\newcommand{\specialcell}[2][l]{%
  \begin{tabular}[#1]{@{}l@{}}#2\end{tabular}}

\usepackage[pagebackref=true,breaklinks=true,letterpaper=true,colorlinks,bookmarks=false]{hyperref}

\usepackage{setspace}
\usepackage{enumitem}
\begin{document}

\title{Semantic Segmentation from Remote Sensor Data and the Exploitation of Latent Learning for Classification of Auxiliary Tasks}
\author{Bodhiswatta Chatterjee,~\IEEEmembership{Student Member,~IEEE} Charalambos~Poullis,~\IEEEmembership{Senior Member,~IEEE}
\IEEEcompsocitemizethanks{
\IEEEcompsocthanksitem B. Chatterjee is a researcher at the Immersive and Creative Technologies lab and an Applied Scientist at Presagis Inc, Canada.\protect\\
E-mail: bodhiswattachatterjee@gmail.com\protect\\
\IEEEcompsocthanksitem C. Poullis is the Director of the Immersive and Creative Technologies Lab at the Department of Computer Science and Software Engineering, Concodia University, Montreal, Quebec, Canada.  \protect\\
E-mail: charalambos@poullis.org - see http://www.poullis.org 
}
\thanks{Manuscript received April 19, 2005; revised August 26, 2015.}}

\markboth{B. Chatterjee, C. Poullis: Semantic Segmentation and Latent Learning for Classification of Auxiliary Tasks}{}
%

\IEEEtitleabstractindextext{%
\begin{abstract}

In this paper we address three different aspects of semantic segmentation from remote sensor data using deep neural networks. Firstly, we focus on the semantic segmentation of buildings from remote sensor data and propose ICT-Net: a novel network with the underlying architecture of a fully convolutional network, infused with feature re-calibrated Dense blocks at each layer. Uniquely, the proposed network combines the localization accuracy and use of context of the U-Net network architecture, the compact internal representations and reduced feature redundancy of the Dense blocks, and the dynamic channel-wise feature re-weighting of the Squeeze-and-Excitation(SE) blocks. The proposed network has been tested on the INRIA and AIRS benchmark datasets and is shown to outperform all other state of the art by more than 1.5\% and 1.8\% on the Jaccard index, respectively. 

Secondly, as the building classification is typically the first step of the reconstruction process, we investigate the relationship of the classification accuracy to the reconstruction accuracy. A comparative quantitative analysis of reconstruction accuracies corresponding to different classification accuracies confirms the strong correlation between the two. We present the results which show a consistent and considerable reduction in the reconstruction accuracy. 

Finally, we present the simple yet compelling concept of \textit{latent learning} and the implications it carries within the context of deep learning. We posit that a network trained on a primary task (i.e. building classification) is unintentionally learning about auxiliary tasks (e.g. the classification of road, tree, etc) which are complementary to the primary task. Although embedded in a trained network, this latent knowledge relating to the auxiliary tasks is never externalized or immediately expressed but instead only knowledge relating to the primary task is ever output by the network. We experimentally prove this occurrence of incidental learning on the pre-trained ICT-Net and show how sub-classification of the negative label is possible \textit{without further training/fine-tuning}. We present the results of our experiments and explain how knowledge about auxiliary and complementary tasks - for which the network was never trained - can be retrieved and utilized for further classification. We extensively tested the proposed technique on the ISPRS benchmark dataset which contains multi-label ground truth, and report an average classification accuracy (F1 score) of 54.29\% (SD=17.03) for roads, 10.15\% (SD=2.54) for cars, 24.11\% (SD=5.25) for trees, 42.74\% (SD=6.62) for low vegetation, and 18.30\% (SD=16.08) for clutter.

The source code and supplemental material is publicly available at \url{http://www.theICTlab.org/lp/2019ICT-Net/}.
\end{abstract}

}

\maketitle

\IEEEdisplaynontitleabstractindextext

%
\IEEEpeerreviewmaketitle

\IEEEraisesectionheading{\section{Introduction}\label{sec:introduction}}

%
%
%
%
\IEEEPARstart{I}{n} recent years there have been tremendous advances in the area of computer vision due to the advent of deep learning. Human-like performance - and in some cases superhuman as in \cite{szegedy2017inception} - has already been achieved for the cognitive task of visual and spatial processing. However, this high performance is often limited to test data which exhibits similar characteristics to the training dataset e.g. has the same feature space with the same distribution. Thus, in order for a network to achieve high accuracy it is imperative that it is trained on a large dataset having large variability.

To address this need, many online competitions are offering large benchmark datasets for training, validation, and testing. The creation of a large training dataset involving images or video requires an extensive number of people to manually label the data in order to ensure the correctness and completeness of the ground truth. This is exacerbated by the fact that each dataset typically serves a single classification task. This is especially true for remote sensor data. For example, in our work we use two datasets provided by INRIA \cite{maggiori2017dataset} and AIRS \cite{DBLP:journals/corr/abs-1807-09532} which include only positive labels for buildings, therefore can only be used for this specific binary classification task. 

Despite the fact that to create such datasets is a tedious, time consuming, and often very expensive process, researchers are relying on large datasets for training classifiers to assist in the solution of more difficult problems, such as reconstruction. More specifically, reconstructing large-scale urban areas is an inherently complex problem which involves a number of vision tasks. Typically, the first step is classification where the objective is to label each pixel into an urban feature type e.g., building, road, tree, car, ground, vegetation, etc. Next, the pixel-level labels are used to cluster the pixels into contiguous groups corresponding to instances of the urban features they represent. Finally, the reconstruction is performed on each cluster. A reconstruction algorithm is applied on each cluster according to the urban feature type the cluster corresponds to. In the case of clusters corresponding to buildings, a boundary refinement process is typically performed prior to extruding the building facades.  

Hence, as it is evident from the pipeline described above, classification is an imperative component of the process and if the objective is to achieve a complete urban-area reconstruction one has to first address the problems relating to the classification. Thus, in this paper our contributions relate to the following three aspects of classification - more specifically semantic segmentation - from remote sensor data using deep neural networks:
\begin{itemize}[leftmargin=*]

\item Firstly, we address the problem of the classification of buildings in remote sensor imagery. We investigate a number of state of the art deep neural network architectures and present a comparative study of the results along with a reasoned justification on the design decisions for the proposed network named ICT-Net: a novel network with the underlying architecture of a fully convolutional network infused with Dense feature re-calibrated blocks at each layer. We demonstrate that this combination of components leads to superior performance. The proposed network is ranked first \footnote{As of February 2019} at two international benchmark competitions organized by INRIA and AIRS, with higher performance by $1.5\%$ and $1.8\%$ respectively from the second best networks.

\item Secondly, we study the relationship between the classification accuracy and reconstruction accuracy. We perform a comparative quantitative analysis on the reconstructions corresponding to classifications of different accuracies and report the results. Due to the lack of depth information, reconstructing 3D models is not feasible therefore the accuracy of the border localization is used as proxy for the evaluation since it is tightly coupled to the reconstruction accuracy i.e. buildings are extruded using their boundaries. As anticipated there is a strong correlation between the classification accuracy and the accuracy of the reconstruction however the analysis has shown that there is a consistent and considerable decrease in the reconstruction accuracy in terms of the per-pixel and per-building Jaccard indices. To the best of our knowledge this is the first time a quantitative analysis is performed in order to establish how the classification accuracy relates to the accuracy of the reconstruction as determined by the accuracy of the border localization.

\item Thirdly, we introduce the concept of \textit{latent learning}  in networks trained with datasets tailored for a binary classification task. In psychology, latent learning (or incidental learning) is a form of learning which occurs unintentionally \cite{tolman1930introduction}. It stems from our ability as humans to learn without making any conscious attempt and thus retain information not directly relevant to the primary task we are performing, without any reinforcement or motivation i.e. no reward or penalty. 

This concept can be transferred to the context of learning in deep neural networks. It would therefore mean that at the same time a deep neural network is trained to perform a primary task for which labeled data is required, latent learning is also occurring for auxiliary tasks for which labeled data is not required. In other words, supervised learning is performed on a primary task while unsupervised learning is concurrently occurring on auxiliary tasks. 

We experimentally prove that latent learning is indeed occurring in the training of deep neural networks, and demonstrate how the latent knowledge can be externalized on the pre-trained binary classification convolutional neural network ICT-Net. Although it has been reported in literature that clustering of features relating to \textit{labeled classes} occurs throughout the network for simple architectures e.g.  AlexNet\cite{AlexNet}, VGG \cite{AlexNet}, to the best of our knowledge this is the first time this is shown for semantic segmentation networks with very large depth, for classes \textit{not} contained in the training labels. Our experiments indicate that deep neural networks encode and cluster information that is not directly relevant to the primary task thus making it possible to re-purpose a pre-trained network for sub-classification of auxiliary tasks without further training or fine-tuning. The proposed technique was evaluated on a third benchmark dataset (ISPRS) which contained multi-label ground truth for 6 classes: buildings, roads, cars, trees, low vegetation, and clutter. Using the proposed technique and without further training, we are able to achieve accuracies of 54.29\% (SD=17.03), 10.15\% (SD=2.54), 24.11\% (SD=5.25), 42.74\% (SD=6.62), 18.30\% (SD=16.08) for auxiliary tasks relating to the classification of roads, cars, trees, low vegetation, and clutter respectively.
\end{itemize}

\noindent
\textbf{Paper organization:} The paper is organized as follows: Section \ref{sec:related_work} presents an overview of state of the art in the area of classification of urban features in satellite images using deep neural networks, and in the area of interpretability of neural networks. Section \ref{sec:system_overview} provides an overview of the work. The proposed neural network ICT-Net trained on the binary classification task of building/non-building is explained in Section \ref{sec:building_classification} including a reasoned justification of the design decisions, and details on the training and testing of the network. Section \ref{sec:reconstruction_analysis} presents a quantitative analysis of the reconstruction accuracies resulting from different classification accuracies. Section \ref{sec:latent_learning} describes the hypothesis and experimental proof of latent learning and explains how a pre-trained network can be re-purposed for classification of auxiliary tasks by sub-classification of the negative label. Finally, Section \ref{sec:conclusion} concludes the work and discusses future directions.

\section{Related Work}
\label{sec:related_work}
Below we provide an overview of the most relevant work relating to (a) the classification  of  urban  features  from remote sensor data using deep learning,  and (b) interpretability in deep neural networks.

\subsection{Classification}
Over the years object recognition has become one of the most addressed vision challenges and as a result a plethora of work has already been proposed. Initially the goal was on image classification where the entire image was classified according to the single object it contained; with some of the most important work in this area being \cite{AlexNet, VGGNet, InceptionNet}. More recently, the objective has shifted towards the semantic object segmentation or semantic labeling where multiple objects contained in a single image are labeled according to their class at the pixel level. Some of the most important work in this area is the work in \cite{FCN, SegNet, UNet}. Recent techniques using deep neural networks have demonstrated excellent results. Below we provide a brief overview of the state of the art related to the area of semantic labeling with a particular focus on remote sensor imagery. A comprehensive review of neural network architectures for semantic segmentation can be found in \cite{SemanticSegmentationReview}.



Typical semantic segmentation architectures comprise of a down-sampling path responsible for feature extraction and an up-sampling path to restore the resolution of the semantic labels. Skip connections between the two paths help to have a smooth gradient back-propagation and fast training of the network. The U-Net \cite{UNet} architecture was able to achieve end-to-end semantic labeling with high accuracy in the field of medical image segmentation. Since then the U-Net \cite{UNet} architecture has been extensively used and adapted to many other domains especially labeling of buildings from aerial imagery as in \cite{StackedUNet, MultiTaskUNet, TernusNetV2}.

At the same time, deeper networks \cite{InceptionNet} have demonstrated the capacity to extract better features from images. Skip connections have been shown \cite{ResNet} to play a critical role in the training of very deep networks as they facilitate very good gradient propagation. There has been a lot of work on the pattern of skip connections with a very promising pattern known as Dense blocks proposed in \cite{DenseNet} for the problem of image classification. In a Dense block every layer is connected to every other layer in a feed-forward fashion. This provides implicit deep supervision and feature reuse which in turn improves the feature extraction power without making it difficult to train the network. The Tiramisu network architecture proposed in \cite{TiramisuNet} extended the use of Dense blocks for semantic segmentation and was able to outperform state of the art on two benchmark data sets: Gatech and CamVid.

Most deep neural networks for object recognition consider all extracted features at each layer to be of equal importance. This was until the method proposed in \cite{SENet} showed that feature re-calibration i.e. weighing of the features, can be used effectively to model inter-dependencies between channels and produce even better performance with little computational overhead. Along a similar direction, in \cite{FCNSE} the authors have shown that feature re-calibration combined with well known FCN networks perform well for medical image segmentation. 

With respect to urban reconstruction, the extraction of urban geospatial features such as buildings from remote sensor imagery has also been an area of research interest for a very long time \cite{shackelford2004automated, wang2006bayesian, haithcoat2001building}. Automatic reconstruction of 3D models from the extracted features is extremely useful for many applications ranging from urban and community planning, development and architectural design, training of emergency response personnel, military personnel, etc. In \cite{5206562} the authors propose a novel, robust, automatic segmentation technique based on the statistical analysis of the geometric properties of the data as well as an efficient and automatic modeling pipeline for the reconstruction of large-scale areas containing several thousands of buildings. With the recent advances in deep neural network architectures the pipeline has been upgraded to feature extraction using a semantic labeling CNN followed by clustering the points based on their label, and specialized processing for each of the labels of geospatial objects as proposed in \cite{Forbes2018DeepAW}.

Recently there has been a lot of interest for semantic labeling of buildings \cite{mnih2013machine, TernusNetV2, StackedUNet} fueled by the release of very large datasets such as the INRIA Aerial Image Labeling dataset \cite{maggiori2017dataset}, and SpaceNet where a corpus of commercial satellite imagery with labeled training data was made publicly available for use in machine learning research. In \cite{TernusNetV2} the authors use a variant of the aforementioned U-Net network architectures replacing the VGG11 \cite{VGGNet} encoder with a more powerful activated Batch Normalized \cite{DBLP:journals/corr/abs-1712-02616} WideResnet-38 \cite{DBLP:journals/corr/ZagoruykoK16} in the context of instance segmentation of buildings for DeepGlobe-CVPR 2018 building detection sub-challenge, and were able to get very good results. 

In this work, we propose ICT-Net: a novel network architecture that combines the strengths of deep neural network architectures (UNet) and building blocks (DenseNet block, SE block) which when applied to the problem of semantic labeling of buildings is proven to achieve better classification accuracy than state of the art on the INRIA Aerial Image Labeling dataset. As of writing this manuscript the proposed network is top ranked since February 2019 on 2 benchmark competitions' leaderboard with more than 1.5\% and 1.8\% difference from the second best entry.

\begin{figure*}[!ht]
    \centering
    \includegraphics[width=\textwidth]{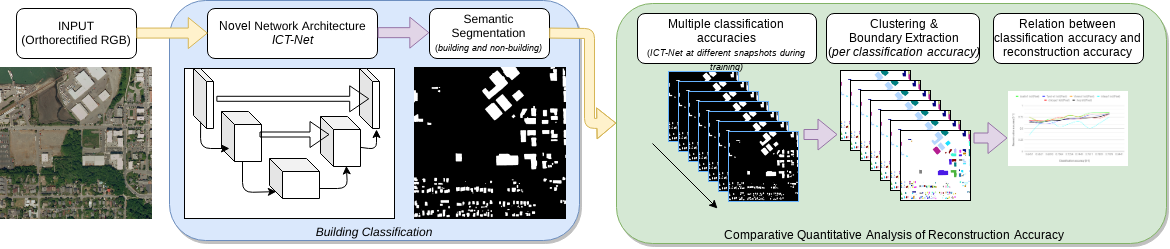}
    \caption{The diagram summarizes the work presented in this paper. Firstly, we focus on the building classification and propose a novel network architecture which outperforms state-of-the-art on benchmark datasets and is currently top-ranking. Secondly, we investigate the relation between the classification accuracy and the reconstruction accuracy and conduct a comparative quantitative analysis which shows a strong correlation but also a consistent and considerable decrease of the reconstruction accuracy when compared to the classification accuracy.}
    \label{fig:system_overview}
\end{figure*}

\subsection{Interpretability}

As soon as the deep neural networks started performing well there has been a huge emphasis on understanding the internal workings of deep neural networks. Zeiler et al.  \cite{DBLP:journals/corr/ZeilerF13} proposed deconvolutional networks to visualize and improve the performance of Convolutional Neural network like AlexNet\cite{AlexNet}. After that there has been a lot of effort to interpret the internal workings of deep neural networks. All currently known techniques focus on interpretibility of image classification networks. These techniques can be roughly categorised into 4 different groups (i) Saliency based techniques like \cite{simonyan2013deep, selvaraju2017grad, sundararajan2017axiomatic} investigate change of pixels which lead to increase or decrease of prediction probability for some specific class of objects, (ii) Decomposition based techniques like \cite{bach2015pixel, zhang2018top, montavon2017explaining} decompose the prediction of the network output into the relevance scores and propagate back to the previous layer, ultimately leading back to the input image to find which input pixels are responsible for a prediction, (iii) Deconvolutional approaches like \cite{DBLP:journals/corr/ZeilerF13, springenberg2014striving} try to map lower layer activations to higher layers and ultimately back to the input image to have a better understanding of what kind of patches provide some specific layer activations, (iv) Another important line of work proposed in \cite{mahendran2015understanding, zhou2016learning, bau2017network, fong2018net2vec} is understanding the predictions of the model by analyzing the individual units of the network and trying to come up with an explanation for the most common patches that produce the specific activation.

The discussed techniques provide a very brief overview of the huge amount of work done in the recent past on interpreting deep neural networks. All the currently known techniques focus on interpreting classification networks where there is only one label for the complete image. There are other tasks in computer vision like semantic segmentation where neural networks based architectures are performing very well but there has been very little focus on interpreting the network prediction. The currently known techniques for interpreting image classification networks can not be directly applied to image segmentation networks as they have an encoder-decoder architecture as compared to an only encoded architecture of classification networks. In this work we interpret the predictions of our proposed ICT-Net architecture by analysing the last few layers of the network, which to the best of our knowledge is the first time to be done in the context of semantic segmentation.

Perhaps the closest concept to what we are proposing is that of auxiliary learning where in order to improve the ability of a primary task to generalize to new unseen data, an auxiliary task is trained in unison. This, of course, has the requirement that labels for the auxiliary task are also available for supervised learning as in \cite{liebel2018auxiliary}. Hence, an alternative would be unsupervised auxiliary learning similar to the work presented in \cite{flynn2016deepstereo, jaderberg2017reinforcement, zhang2018fine}. 
Although these methods eliminate the need for labeled data, they typically involve a secondary label generation network which is tightly coupled to the network trained on the primary task as in \cite{liu2019self}. In contrast to the above work and definition of auxiliary learning, our work refers to the \textit{unintentional learning of auxiliary tasks} resulting from the learning of a primary task for which labeled data is available and easily obtainable.

In summary, previous work has only dealt with the related yet different concepts of interpretability of neural networks, and auxiliary tasks. In contrast, our technical contribution is on exploiting latent learning in order to re-purpose a pre-trained classification network on auxiliary tasks for which labeled data is not available. To the best of our knowledge, this is the first time the concept of latent learning is introduced within the context of deep neural networks.

\section{System Overview}
\label{sec:system_overview}
Figure \ref{fig:system_overview} gives an overview of our first two contributions. Firstly, an orthorectified RGB image is fed forward into the neural network to produce a binary (building/non-building) classification map. Next, the binary classification map becomes the input to the reconstruction process. Due to the fact that it is extremely difficult to acquire building blueprints or CAD models for such large areas and depth/3D information is not available for the images of the benchmark, we posit that the building boundaries extracted from the binary classification map and subsequently refined, can serve as a proxy to the accuracy of the reconstruction. This is justified since the extracted boundaries are extruded in order to create the 3D models for the buildings. Therefore, the building boundaries are extracted, refined, and are used in the comparative analysis and evaluation of the accuracy of the reconstruction.




\section{Building Classification}
\label{sec:building_classification}
In this section we describe the details of the proposed neural network architecture including information about the datasets, training/validation, and testing, as well as the justification for all design decisions.

\subsection{Datasets}
\label{subsec:dataset}

We have chosen the two datasets described below over other available options because they uniquely offer two significant advantages. Firstly, the training and testing datasets are from \textit{completely different cities} with no overlap.
Secondly, the datasets cover \textit{dissimilar urban settlements} e.g., European, American, etc, with large variability in building density, architecture, and overall characteristics e.g., red shingles, flat roofs, etc. For these reasons, we have chosen the following benchmark datasets because they are ideal for assessing the \textit{generalization capacity of the network}.

\subsubsection{INRIA Aerial Image labeling dataset}
The training of the network is performed using the INRIA Aerial Image labeling dataset \cite{maggiori2017dataset} which consists of pixelwise labeled aerial imagery for building classification. The dataset covers $810 km^2$ area across 10 different cities with ground sampling density resolution of $30cm$, and is split into two equal sets ($405 Km^2$ each) for training and testing. The dataset consists of 3-band orthorectified RGB images and the training labels consist of ground truth data for two semantic classes: building and non-building. The training data covers parts of the cities of Austin, Chicago, Kitsap county, western Tyrol, and Vienna. The test data covers parts of the cities of Bellingham, Bloomington, Innsbruck, San Francisco and Eastern Tyrol. There are 36 tiles with resolution of $5000 \times 5000$ pixels for each city, each tile covering $1500 \times 1500m^{2}$  area on the ground. The training data is further divided into two sets: (1) the validation set which comprises of the first 5 tiles of each city, and (2) the training set which consists of the rest of the tiles as suggested in \cite{maggiori2017dataset}. An example image from the dataset can be seen in Figure \ref{fig:system_overview}. 

\subsubsection{AIRS dataset}
AIRS (Aerial Imagery for Roof Segmentation) is a public dataset that aims at benchmarking the algorithms of roof segmentation from very high-resolution aerial imagery. AIRS dataset covers almost the full area of Christchurch, the largest city in the South Island of New Zealand. Although the aerial imagery is from one city, there is huge variety of settlement types. The imagery contains 3-band orthorectified RGB images at $7.5cm$ ground sampling density. It has a coverage of $457 Km^2$ aerial images with approximately over 220,000 buildings and refined ground truths that strictly align with roof outlines.There are 1046 tiles with resolution of $10000 \times 10000$ pixels which are already divided into train, validation and test sets of sizes 857, 94, and 95. The ground truth for buildings is carefully refined to align with their roofs and the segmentation task for AIRS contains two semantic classes: roof and non-roof. We have chosen the AIRS benchmark dataset as it covers a completely different geographic location with different urban settlements and the aerial imagery is very high resolution so we are able to further validate the \textit{generalization capacity} of the trained neural network.

\subsection{Network Architecture}
\label{subsec:network_architecture}

A vast number of networks has been proposed for image classification and semantic labeling. State of the art performance is generally achieved with deep networks however these are difficult to train due to vanishing or exploding gradients. Many networks \cite{ResNet, DenseNet, UNet, TiramisuNet} have shown skip connections play an important role in having good gradient propagation through the network. In our work, as part of the network design process, we first identified the requirements for the particular task at hand i.e. semantic segmentation of buildings from remote sensor images, and then decisions were made to address these:

\begin{itemize}
    \item \textbf{Requirement 1:} An important aspect of semantic segmentation of buildings is to have high localization accuracy and take into account as much context information as possible. This is necessary in order to address the wide variability in buildings typically relating to their function e.g., shape, size, color and/or region they appear in e.g., density in urban/rural, etc. 
    
    \textbf{Decision:} To that end, the U-Net architecture \cite{UNet} takes into account spatial information and combines it with contextual information via the direct downsampling-upsampling links. 
    
    \item  \textbf{Requirement 2:} In order to be able to process large chunks of data at a time it is imperative that the network contains as few parameters as possible. 
    
    \textbf{Decision:} Dense blocks connect every layer to every other layer in a feed-forward fashion. Along with good gradient propagation they also encourage feature reuse and reduce the number of parameters substantially as there is no need to relearn the redundant feature maps. At the end of every Dense block all the extracted features accumulate, creating a very diverse set of features. As a result of this feature redundancy there is a substantial reduction in the network parameters leading to faster training times. This allows the processing of larger patch (and batch) sizes (which also addresses Requirement 1) therefore allowing additional contextual information during each feed-forward pass.


    \item  \textbf{Requirement 3:} The contribution of the feature maps at each layer to the output must depend on their importance. 
    
    \textbf{Decision: } Using the Squeeze-and-Excitation (SE) blocks the dynamic channel-wise feature re-weighting mechanism provides a way to upweigh important feature maps and downweigh the rest. In \cite{SENet} authors show adaptive re-calibration of channel-wise feature responses by explicitly modelling inter-dependencies between channels using squeeze and excitation block on existing architectures \cite{ResNet, InceptionNet, ResNext} results in improved performance. 

\end{itemize}

\begin{figure}
    \includegraphics[width=0.5\textwidth]{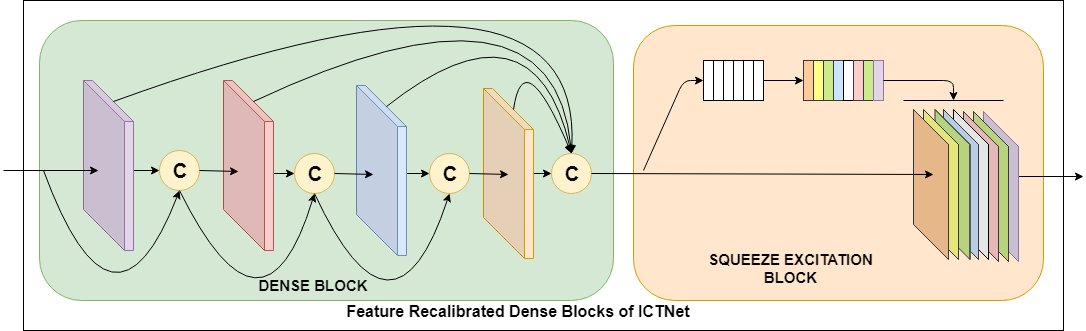}
    \caption{Proposed feature recalibrated Dense block with 4 convolutional layers and a growth rate $\kappa = 12$ used by the ICT-Net. c stands for concatenation.}
    \label{fig:network_architecture}
\end{figure}

The proposed network architecture is distinct and combines the strengths of the U-Net architecture, Dense blocks, and Squeeze-and-Excitation (SE) blocks. This results in improved prediction accuracy and it has been shown to outperform other state of the art network architectures such as the ones proposed in \cite{huang2018large} which have a much higher number of learning parameters on the INRIA benchmark dataset. Figure  \ref{fig:network_architecture} shows a diagram of the proposed feature recalibrated Dense block with 4 convolutional layers and a growth rate $\kappa = 16$ used by the ICT-Net. The proposed network has 11 feature recalibrated dense blocks with [4,5,7,10,12,15,12,10,7,5,4] number of convolutional layers in each dense block respectively.

Perhaps the closest architecture to the one proposed was discussed in \cite{TiramisuNet} which uses 103 convolutional layers. If SE blocks are introduced at the output of every layer this will cause a vast increase in the number of parameters which will hinder the training. In contrast, in our work we have chosen to include an SE block only at the end of every Dense block in order to re-calibrate the accumulated feature-maps of all preceding layers. Thus, the variations in the information learned at each layer - in the form of the features maps - are weighted by the SE block according to their importance as determined by the loss function. 


\noindent
\textbf{Discussion: } To verify the validity of the above design decisions we performed a comparative study involving a number of state of the art architectures and blocks. Following the same training procedure for all architectures reported, and without any data augmentation the ICT-Net was compared with U-Net \cite{UNet} and Tiramisu-103 \cite{TiramisuNet}. The results on the validation dataset are shown in Table \ref{tab:validation_comparison_table} where it is evident that the proposed architecture outperforms both U-Net and Tiramisu-103.


\begin{table}[!ht]
    \centering
  \begin{tabular}{|| l | l | c | p{44pt} || }
    \hline
    Paper & Method & Overall IoU (\%) & Overall Accuracy (\%) \\ \hline
    \cite{UNet} & UNet & 70.86 & 95.51 \\ \hline
    \cite{TiramisuNet} & Tiramisu-103 & 73.91 & 95.71 \\ \hline
    Ours & ICT-Net & \textbf{75.5}  & \textbf{96.05} \\ \hline
    \hline
  \end{tabular}
  \caption{Performance evaluation of SOTA architectures (U-Net \cite{UNet} and Tiramisu-103 \cite{TiramisuNet}) on the validation dataset}
  \label{tab:validation_comparison_table}
\end{table}

\subsection{Training and Validation on INRIA dataset}
The network is trained on 155 tiles each with resolution $5000 \times 5000$ from the available training data with their corresponding ground truth. The training is performed for 100 epochs on a single nVidiaGTX 1080Ti. We used Tensorflow API for the development and training/testing of the network. Due to the large size of the dataset it requires approximately 6 hours to complete 1 epoch of training. Every epoch was divided into 31 sub-epochs each consisting of 5 tiles (1 from each city). Limited by GPU memory we had to choose a small batch size of 4 to have a comparatively larger patch size of $256 \times 256$ as we observed context is very important for semantic labeling of buildings.

\textbf{Implementation details: } The network was trained using cross-entropy loss with RMSProp Optimizer with an initial learning rate of 0.001 and decay of 0.995 for the first 50 epochs. After the $50^{th}$ epoch the learning rate was reduced to 0.0001 and trained for another 50 epochs. Instead of using dropout as a regularization technique we applied a large number of data augmentations in order to restrict the network from overfitting to the training dataset. 

\textbf{Data input:} Our network takes in patches of $256\times256$ out of the entire tile with 50\% overlap. The patches are selected sequentially for every odd epoch and the same number of patches is selected randomly for every even epoch during the training. We use the alternating patch generation strategy to restrict the network from overfitting while still having the opportunity to learn all the features from every tile. At testing the input patch size in increased to $768\times768$ (the maximum that could fit in the GPU memory) so that we are able to increase the context for large buildings in every patch. During testing, the patches are selected using 50\% overlap similar to what is done during training.

\begin{figure}
    \centering
    \includegraphics[width=0.48\textwidth]{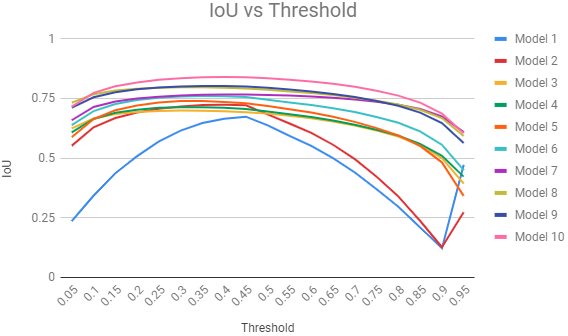}
    \caption{Empirical study to determine the optimal thresholding value for converting the grayscale classification map produced by the network to a binary map. The models shown correspond to the same network ICT-Net at different training snapshots for which the classification accuracy (i.e. IoU in the graph) was calculated \textbf{after} the thresholding at every $0.05$ intervals as shown. The optimal threshold value is $\tau=0.4$.}
    \label{fig:thresholding}
\end{figure}

\textbf{Network output: } The output produced by the network is a 1-channel gray-scale image of the same size as the input image where each pixel has a probability score of being a building in the range $[0,1]$. We convert the probability map into a binary mask by thresholding. We conducted an empirical study on the validation dataset and have chosen $\tau=0.4$ as the optimal threshold value for converting the gray-scale image to a binary map as shown in Figure \ref{fig:thresholding}. The output patches are then assembled into tiles of size $5000\times5000$ by weighted average and overlapping areas near the edges are down-weighted. During the testing, the standard test time augmentations are applied to each tile and they are merged back using an average of the probability scores.

\textbf{Data augmentations: } Based on the validation results we used the pre-trained weights and trained our network with the following data augmentations with a probability of 70\% to be applied to every patch: random rotations in the range $[0^{\circ}, 360^{\circ}]$ using reflection padding, random flip, random selection of a patch in the range of $[0.75, 1.25]$ of the image patch size and re-size it to original patch size of 256. Data augmentations significantly improved the performance of the network in terms of accuracy.

\subsection{Training and Validation on AIRS dataset: }
Initially we use the pre-trained ICT-Net trained on INRIA dataset to test the generalization ability of the network on AIRS validation dataset of 94 tiles. Then the network is trained on 857 tiles of AIRS training dataset where each tile is of resolution $10000 \times 10000$ with their corresponding ground truth. The training is performed for 5 epochs on a single nVidiaGTX 1080Ti. Due to the large size of the dataset it requires approximately 30 hours to complete 1 epoch of training. Every epoch was divided into sub-epochs where each sub-epoch consists of 5 tiles from the training dataset. We use the same batch and patch size of 4 and $256 \times 256$ respectively as we used for INRIA dataset. The optimization hyper-parameters remain the same and the learning rate used was 0.0001 for all 5 epochs. The output patches produced by the network are then assembled into tiles of size $10000\times10000$ by weighted average and overlapping areas near the edges are down-weighted. On the AIRS dataset we use the same set of data augmentation techniques that were used while training ICT-Net on INRIA dataset.

\subsection{Evaluation - INRIA Test dataset}
The INRIA dataset uses two main performance measures: Intersection over Union (Jaccard index) and Accuracy. Intersection over Union (IoU) is defined as the number of pixels labeled as buildings in both the prediction and the reference, divided by the number of pixels labeled as buildings in the prediction or the reference. Accuracy is defined as the percentage of correctly classified pixels.

\begin{figure}[!ht]
    \centering
    \includegraphics[width=0.5\textwidth]{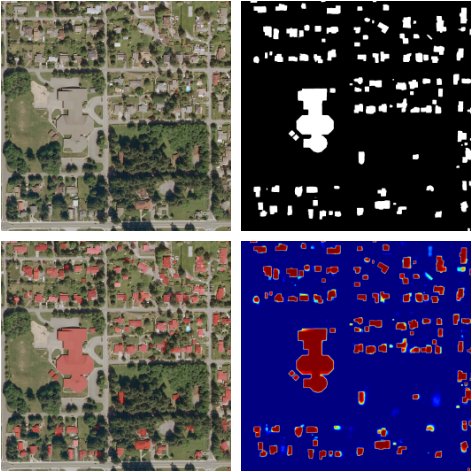}
    \caption{Building Classification of Bellingham city, tile 17. Result for an image from the test dataset.  Top left: A closeup of a small area of an input image. Bottom left: The probability image overlaid on top of the RGB image. Bottom right: The probability image shown as a heat map. Top right: The binary map resulting after the thresholding of the probability image.}
    \label{fig:heat_map1}
\end{figure}

The measures are calculated by the organizers of the competition and involve the classification of 5 cities for which no images have been used for training and validation, and for which no ground truth is available to the participants. As of writing this manuscript the proposed architecture is ranked as the top performing since February 2019 in terms of both IoU (80.32\%) and accuracy (97.14\%) on the competition's leaderboard \footnote{\url{https://project.inria.fr/aerialimagelabeling/leaderboard/}}. Figure \ref{fig:heat_map1} shows an example of a result for a small area of an image from the test dataset (top left). The probability image produced by the network is shown as a heat map (bottom right) overlaid on top of the RGB image (bottom left). The binary map resulting after the thresholding is shown in the top right image.

As previously mentioned, the proposed network is currently ranked as the top performing network with the second best having more than $1.5\%$ difference in terms of the IoU. The competition organizers in \cite{huang2018large} provide details of the next 4 top performing techniques on the INRIA aerial image labeling benchmark dataset. All 4 methods are Convolutional Neural Networks(CNNs), among which 3 of them are based on U-Net architecture. Table \ref{tab:comparison_table} shows a quantitative comparison between the proposed network ICT-Net and these other techniques on the test dataset as reported by the competition organizers. 


\begin{table}[!ht]
    \centering
  \begin{tabular}{|| l | l | c | c || }
    \hline
    Paper & Method & Overall IoU & Overall Accuracy \\ \hline
    \cite{huang2018large} & Raisa & 69.57  & 95.30 \\ \hline
    \cite{huang2018large} & ONERA & 71.02  & 95.63 \\ \hline
    \cite{huang2018large} & NUS & 72.45 & 95.90 \\ \hline
    \cite{huang2018large} & AMLL & 72.55 & 95.91 \\ \hline
    \cite{INRIAleaderboard} & N/A & 78.31 & 96.76 \\ \hline
    \cite{INRIAleaderboard} & N/A & 78.39 & 96.84 \\ \hline
    \cite{INRIAleaderboard} & N/A & 78.45 & 96.74 \\ \hline
    \cite{INRIAleaderboard} & N/A & 78.80 & 96.91 \\ \hline
    \hline Ours & ICT-Net & \textbf{80.32} & \textbf{97.14} \\
    \hline
  \end{tabular}
  \caption{Performance evaluation of the best performing networks on the test dataset. First 4 entries are defined as State of the art by INRIA\cite{huang2018large}. Next 4 entries are other top performances from the leaderboard\cite{INRIAleaderboard}. ICT-Net outperforms all others with more than $1.5\%$ difference in terms of the IoU.}
  \label{tab:comparison_table}
\end{table}

\begin{figure*}[!ht]
    \centering
    \begin{subfigure}[b]{0.49\textwidth}
            \includegraphics[width=\textwidth]{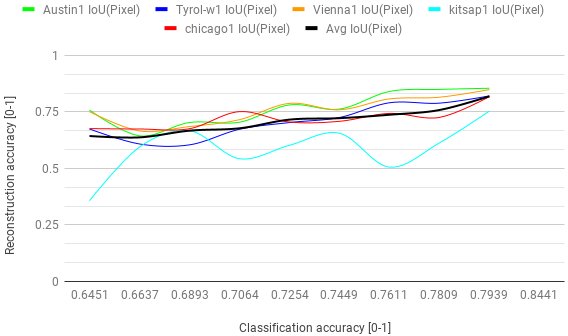}
            \caption{}
            \label{fig:reconstruction_vs_classification1}
    \end{subfigure}
    \begin{subfigure}[b]{0.49\textwidth}
        \includegraphics[width=\textwidth]{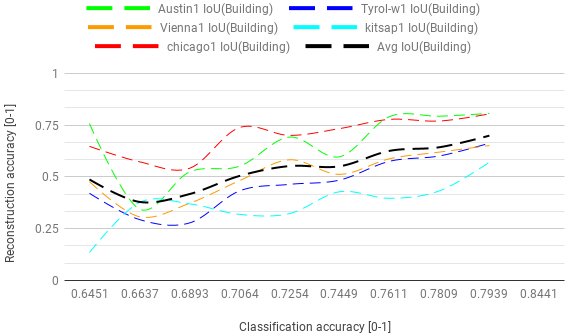}
            \caption{}
            \label{fig:reconstruction_vs_classification2}
    \end{subfigure}
    \caption{(a) Reconstruction vs Classification accuracy. The classification accuracy ranges from [0.6451, 0.8441] as calculated on the validation test. The different classification accuracies correspond to the same architecture (ICT-Net) but at different snapshots during training. The binary classification map produced at each snapshot: (i) is refined using conditional random fields (CRF), (ii) building boundaries are extracted, (iii) building boundaries are refined using the Douglas-Pecker algorithm, (iv) converted back to a binary classification map, and (v) compared with the ground truth. Two metrics are used: per-pixel IoU and per-building IoU (with a threshold of 75\% overlap for true positive). There is an average decrease of 4.43\% $\pm$ 1.65\% (confidence level 95\%) in per-pixel IoU (a) of the reconstruction accuracy; and an average decrease of 21.7\%$\pm$ 4.21\% (confidence level 95\%) in per-building IoU (b) of the reconstruction accuracy. The reported averages are calculated across the accuracy levels. }
\end{figure*}

\subsection{Evaluation - AIRS dataset}
The AIRS dataset uses the following performance measures: Intersection over Union (Jaccard index), F1-Score, Precision and Recall. The evaluation metrics of intersection over union (IoU) and F1-score are used to reflect the overall performance of the baseline methods. On the other hand, precision and recall indicate the correctness and completeness of the roof segmentation results respectively. The evaluation provided by the organizers of the competition involves the segmentation of roofs for 95 tiles of imagery for which no images have been used for training and validation, and no ground truth is available to the participants. As of November 2019 the proposed architecture is ranked as the top performing in terms of both IoU (91.70\%) and F1-score (95.70\%) on the competition's leaderboard \footnote{\url{https://www.airs-dataset.com/leaderboard/}}.

The proposed network achieves a $1.8\%$ improvement over the second best network in terms of the IoU and $1\%$ difference in terms of F1-score. The authors in \cite{DBLP:journals/corr/abs-1807-09532} provide details of the top performing techniques on the AIRS (Aerial Imagery for Roof Segmentation) benchmark dataset. All of the methods are Convolutional Neural Networks based approaches(CNNs). Table \ref{tab:airs_result_table} shows a quantitative comparison between the proposed ICT-Net and these other techniques on the test dataset as reported by the competition organizers. 

\begin{table}[!ht]
    \centering
  \begin{tabular}{|| p{18pt} | p{43pt} | c | c | c | c || }
    \hline
    Paper & Method & IoU & F1-score & Precision & Recall \\ \hline
    \cite{lin2017feature} & FPN & 0.882 & 0.937 & \textbf{0.963} & 0.913 \\ \hline
    \cite{DBLP:journals/corr/abs-1807-09532} & FPN+MSFF & 0.888 & 0.941 & 0.958 & 0.924 \\ \hline
    \cite{zhao2017pyramid} & PSP & 0.899 & 0.947 & 0.961 & 0.933 \\ \hline
    \cite{chatterjee2019building} & ICT-Net & \textbf{0.917} & \textbf{0.957} & 0.955 & \textbf{0.959} \\
    \hline
  \end{tabular}
  \caption{Performance evaluation of the top performing networks on the AIRS test dataset. ICT-Net outperforms all others with  $1.8\%$ difference in terms of the IoU and $1\%$ in terms of F1-Score.}
  \label{tab:airs_result_table}
\end{table}

\begin{table*}[!ht]
\begin{adjustwidth}{-1cm}{}
    \centering
  \begin{tabular}{|| p{40pt} | p{30pt} | p{30pt} | p{30pt} | p{30pt} | p{30pt} | p{30pt} | p{30pt} | p{30pt} | p{30pt} | p{30pt} | p{30pt} | p{30pt} || }
    \hline
    \multirow{ 2}{*}{\bfseries \specialcell{Classification \\Accuracy}} &
      \multicolumn{2}{c|}{\bfseries Austin1 IoU} &
      \multicolumn{2}{c|}{\bfseries Tyrol-W1 IoU} &
      \multicolumn{2}{c|}{\bfseries Vienna1 IoU} & 
      \multicolumn{2}{c|}{\bfseries Kitsap1 IoU} & 
      \multicolumn{2}{c|}{\bfseries Chicago1 IoU} & 
      \multicolumn{2}{c|}{\bfseries Average IoU} \\
    
    \cline{2-13}
    {\bfseries } & {\bfseries per-pix.} & {\bfseries per-bldg} & {\bfseries per-pix.} & {\bfseries per-bldg} & {\bfseries per-pix.} & {\bfseries per-bldg} & {\bfseries per-pix.} & {\bfseries per-bldg} & {\bfseries per-pix.} & {\bfseries per-bldg} & {\bfseries per-pix.} & {\bfseries per-bldg}\\
    \hline 0.6451 & 0.7038 & 0.5004 & 0.4683 & 0.1887 & 0.7291 & 0.3880 & 0.1063 & 0.02384 & 0.6445 & 0.4952 & 0.5304 & 0.3192 \\
    \hline 0.6637 & 0.7583 & 0.7583 & 0.6749 & 0.4213 & 0.7514 & 0.4776 & 0.3575 & 0.1354 & 0.6765 & 0.6481 & 0.6437 & 0.4881 \\
    \hline 0.6893 & 0.6443 & 0.3451 & 0.6084 & 0.2944 & 0.6660 & 0.3080 & 0.5949 & 0.3770 & 0.6747 & 0.5734 & 0.6377 & 0.3796 \\
    \hline 0.7064 & 0.7034 & 0.5240 & 0.6046 & 0.2780 & 0.6855 & 0.3728 & 0.6671 & 0.3704 & 0.6760 & 0.5420 & 0.6673 & 0.41745 \\
    \hline 0.7254 & 0.7049 & 0.5520 & 0.6735 & 0.4325 & 0.7160 & 0.4820 & 0.5432 & 0.3194 & 0.7516 & 0.7386 & 0.6778 & 0.5049 \\
    \hline 0.7449 & 0.7812 & 0.6926 & 0.7032 & 0.4643 & 0.7881 & 0.5829 & 0.6026 & 0.3230 & 0.7056 & 0.7011 & 0.7162 & 0.5528 \\
    \hline 0.7611 & 0.7630 & 0.5976 & 0.7256 & 0.4850 & 0.7597 & 0.5128 & 0.6561 & 0.4286 & 0.7088 & 0.7336 & 0.7226 & 0.5515 \\
    \hline 0.7809 & 0.8408 & 0.7914 & 0.7907 & 0.5756 & 0.8078 & 0.5879 & 0.5059 & 0.3973 & 0.7436 & 0.7782 & 0.7378 & 0.6261 \\
    \hline 0.7939 & 0.8498 & 0.7936 & 0.7891 & 0.6016 & 0.8153 & 0.6202 & 0.6131 & 0.4328 & 0.7259 & 0.7703 & 0.7586 & 0.6437 \\
    \hline 0.8441 & 0.8549 & 0.8073 & 0.8212 & 0.6634 & 0.8490 & 0.6519 & 0.7541 & 0.5714 & 0.8179 & 0.8050 & 0.8194 & 0.6998 \\ \hline
  \end{tabular}
  \caption{The ICT-Net at different training snapshots having different classification accuracy vs the reconstruction accuracy measured using two metrics: per-pixel IoU, and per-building IoU (with a threshold of 75\% overlap for true positives)}
  \label{tab:reconstruction_comparison}
  \end{adjustwidth}
\end{table*}

\section{Comparative Quantitative Analysis of Reconstruction Accuracies}
\label{sec:reconstruction_analysis}
As previously stated the objectives and contributions of our work are three-fold. In Section \ref{sec:building_classification} we proposed a novel, top ranking architecture for classifying buildings from remote sensor imagery. This binary classification map is typically used as a first step to the reconstruction process since it allows the application of specialized reconstruction algorithms according to the classified type of the pixels. In this section we focus on the equally important aspect of the relation between the classification accuracy and the accuracy of the reconstruction. Since it is extremely difficult to acquire building blueprints or CAD models for such large areas, and no 3D/depth information is available as part of the benchmark dataset we posit that the building boundaries extracted from the classification binary map can serve as a \textit{proxy} to the quality of the reconstruction since the boundaries are typically extruded in order to create the 3D models corresponding to the buildings. More specifically, the procedure for quantitatively evaluating the accuracy of the reconstruction is as follows:

\noindent
\begin{itemize}
    \item Building boundaries $B_{g}$ are extracted from the ground truth provided as part of the training dataset. 
    
    \item The RGB image corresponding to the ground truth above is used as input to  the ICT-Net. The binary classification map $C_{b}$ resulting from feeding forward the RGB image classifies pixels into buildings and non-buildings.
    
    \item The binary classification map $C_{b}$ is refined $C_{b}^{refined}$ using an MRF-based technique where an energy function is minimized via graph-cut optimization for finding an optimal labeling $f_{p}$ for every pixel $p$ such that $f_{p} \rightarrow l$, where $l$ is the new label. The data term of the energy function of a pixel $p$ with label $l_{p_{i}}$ is defined as, 
    \begin{equation}
    E_{d} = 
    \begin{cases}
        10,& \text{if } f(p_{i}) \neq l_{p_{i}}\\
        0,              & \text{otherwise}
    \end{cases}
    \end{equation}
    The smoothness term of the energy function of two neighbouring pixels $p_{1}$ and $p_{2}$ with labels $l_{p_{1}}$ and $l_{p_{2}}$ respectively is defined as, 
    \begin{equation}
    E_{s} = 
    \begin{cases}
        20,& \text{if } l_{p_{1}} == l_{p_{2}} \text{and }  f(p_{1}) \neq f(p_{2})\\
        0,      & \text{otherwise}
    \end{cases}
    \end{equation}
    The values of $10$ and $20$ in the equations were selected such that smoothness is favored over the observed data. 
    
    \item Building boundaries $B_{b}$ are extracted from the refined classification map $C_{b}^{refined}$. A simplification process i.e. Douglas-Pecker approximation with a tolerance of $\tau=0.5$, is applied to the boundaries. This simplification process is a step applied to the building boundaries prior to extruding the 3D model if 3D/depth information is available \cite{poullis2009automatic}, \cite{poullis2013framework}, \cite{poullis2019large}.  
    
    \item The simplified boundaries $B_{b}^{approx}$ are finally converted back to a binary classification map and quantitatively compared to the ground truth $B_{g}$. This comparison involves IoU metrics on (i) a per-pixel and (ii) a per-building bases. In the case of the per-building IoU metric, a true positive is considered only if a building has at least 75\% of its pixels overlap the pixels of the same building in the ground truth.
\end{itemize}

The procedure described above is followed for all input images with no changes to the values and thresholds used; the only varying condition is the classification accuracy. In our experiments, the input images are processed by the proposed ICT-Net at different training snapshots having different classification accuracies. Thus, multiple binary classification maps were produced each with a different classification accuracy.

Table \ref{tab:reconstruction_comparison} shows the quantitative results of the comparison. A total of 5 cities were processed using the aforementioned procedure. Figures \ref{fig:reconstruction_vs_classification1} and \ref{fig:reconstruction_vs_classification2} show the relation between the reconstruction accuracy with respect to the classification accuracy. We have used increasing classification accuracies based on the same architecture (ICT-Net) at different snapshots during the training. Using the binary classification maps we have followed the procedure which is typical to the reconstruction process. Two metrics have been used to assess the reconstruction accuracy, namely per-pixel IoU and per-buildng IoU (with 75\% threshold for being considered a true positive). As expected, the graph shows a strong correlation between the classification accuracy and the reconstruction accuracy. However the reconstruction accuracy is consistently lower than the classification accuracy by an average of 4.43\% $\pm$ 1.65\% (confidence level 95\%) on the per-pixel IoU and an average of  21.7\%$\pm$ 4.21\% (confidence level 95\%) on the per-building IoU. This discrepancy can be attributed to the fact that the ground truth images used for training the network may contain errors and are in most cases manually created which results in much higher classification accuracy than the reconstruction accuracy. Moreover, the high discrepancy on the per-building IoU can be attributed to the fact that a threshold must be used i.e. 75\%, when calculating the true positives.

The results of this analysis clearly indicate \textit{that high classification accuracy does not translate into high reconstruction accuracy}. More importantly though, the results of the analysis clearly indicate that the reconstruction accuracy must be taken into account as part of the loss function along with the classification accuracy during the training of the network.

\begin{figure}[!ht]
    \centering
    \includegraphics[width=0.5\textwidth]{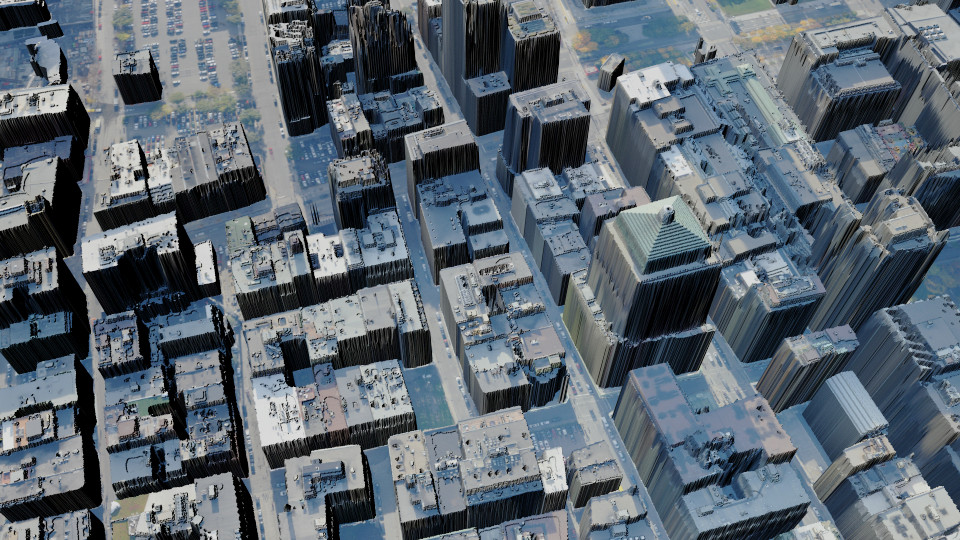}
    \caption{A fully automated result without any post-processing. Downtown Montreal for which no training images were used and no ground truth is available. Classification by ICT-Net and reconstruction by extruding the extracted boundaries of the buildings using the LiDAR pointcloud corresponding to the same area. The elevation of all non-building points is set to zero. All buildings have been manually verified that they are correctly classified. The accuracy of the classification can also be visually verified since there is no "bleeding" between the buildings and any other urban features e.g., roads, trees, cars, etc. The orthophoto RGB image is courtesy of Defence Research and Development Canada and Thales Canada.}
    \label{fig:montreal}
\end{figure}

\begin{figure*}[!ht]
    \centering
    \includegraphics[width=\textwidth]{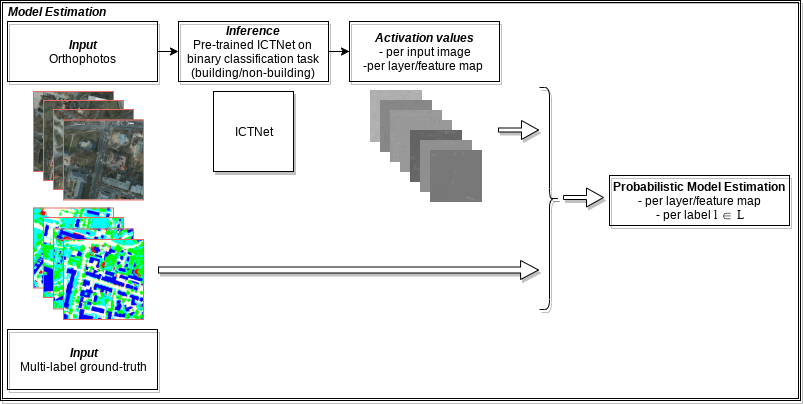}
    \caption{\textbf{Probabilistic model estimation}. We infer the activation values for each feature map at the penultimate layer for four images of size $640\times640$. 
    Multi-label ground-truth corresponding to the four input images is used for aggregate the activation values based on their label. Finally, for each label and for each feature map of the penultimate layer, a probabilistic model $\mathcal{N}(\mu, \sigma$ is estimated. Analysis of the histograms of the activation values shows that it resembles a Gaussian distribution.}
    \label{fig:model_estimation}
\end{figure*}

Figure \ref{fig:montreal} shows an example of the downtown Montreal. The building classification is generated with the proposed ICT-Net network and refined as explained above. In this example, LiDAR information was available which after resampling at the same resolution as the orthorectified image was used to extrude the 3D buildings from the extracted boundaries. The result shown is fully automated and no post-processing was performed. It should be noted that no images of the city of Montreal have been used in the training. We have manually evaluated the result by counting the number of buildings and confirming that all of them have been classified correctly by the network and therefore reconstructed. The accuracy of the classification is also evident from the fact that there is no texture "bleeding" between the buildings and any other urban features e.g., roads, trees, cars, etc in the final result.

\section{Latent Learning in Deep Neural Networks}
\label{sec:latent_learning}

In this section we explain the observations which motivated the hypothesis and how this hypothesis can be proven experimentally. In the following subsections, we use the definition and notation from transfer learning as defined in \cite{pan2009survey} and subsequent literature, on domains and tasks. 

A domain $\Delta$ is defined as $\Delta = \{ \chi^{\Delta}, P^{\Delta}(\textbf{x}) \}$ on feature space $\chi^{\Delta}$ and a marginal probability distribution $P^{\Delta}(\textbf{x})$ over that feature space. $\textbf{x} = x_{1}, x_{2}, \ldots, x_{n} \in \chi^{\Delta}$ are samples from feature space $\chi^{\Delta}$ of the domain $\Delta$. A task $\tau$ consists of a label space $\Upsilon$ from which sample labels $\textbf{y} = y_{1}, y_{2}, \ldots, y_{n} \in \Upsilon$ are drawn. 

In the special case of a binary classification problem the task $\tau_{b}$ is to classify an object as an instance of a class, or otherwise; hence there are only two labels $\{0,1\} \in \textbf{y}$. In this case only the positive label (i.e. $1$) carries important information and $0$ is defined in terms of $1$, and can be simply interpreted as a negative label i.e. $\neg 1$ or "not an instance of the class $1$". This resembles the one-vs-all learning paradigm used for multi-class classification problems which forms the basis for our hypothesis for sub-classification of the negative label $0$.

\begin{figure*}[!ht]
    \centering
    \includegraphics[width=\textwidth]{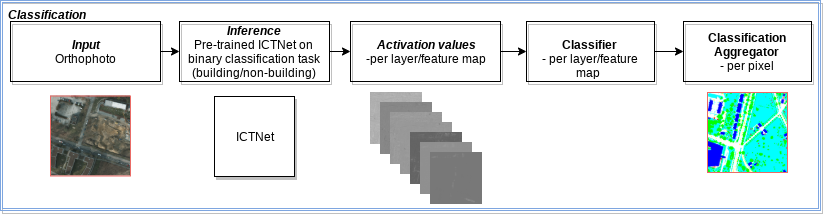}
    \caption{\textbf{Sub-classification of the negative label}. The pre-trained ICT-Net is used to infer the activation values at each feature map of the penultimate layer. A classification image $C_{k}$ is generated for each feature map $k$ using maximum likelihood classification (MLC). Finally, the classification aggregator combines all the $K$ classification images to produce the final classification labels. }
   \label{fig:sub_classification}
\end{figure*}

\subsection{Hypothesis}
\label{subsec:hypothesis}

A neural network trained on domain $\Delta$ with pairs of training samples $(x_{i}, y_{i})$, where $x_{i} \in \textbf{x}$ and $y_{i} \in \textbf{y}$, learns the conditional probability distribution $P_{\Delta}(\textbf{y}| \textbf{x})$. Hence the network can be thought of as approximating an injetive surjective mapping function $f:x_{i} \longrightarrow y_{i}$ from feature space $\chi$ to label space $\Upsilon$. We hypothesize that a neural network trained to perform a task $\tau_{A}$ with labels \textbf{y}$_{\tau_{A}}$, also learns about additional complementary tasks $\tau_{B}, \tau_{C}, \ldots $ where $\{ \tau_{B} \cup \tau_{C} \cup \ldots\} \cap \tau_{A} = \emptyset$, with labels \textbf{y}$_{\tau_{B}}$, \textbf{y}$_{\tau_{C}}$, $\ldots \in \neg$ \textbf{y}$_{\tau_{A}}$.


Intuitively, we posit that in order for a network to learn task $\tau$ with labels $\upsilon$, the network has to learn auxiliary tasks with associated complementary labels $\neg \upsilon$. Thus, one can exploit the latent learning occurring for further sub-classification of the negative label. The above hypothesis is based on the following observations:
\begin{enumerate}
    \item A trained neural network approximates a mapping function from the feature space of a domain to the label space of a task from which the sample labels are drawn.
    
    \item If a pixel is not mapped to a positive label in $\textbf{y}$ then it is assigned the negative label $0$ which can be interpreted as $\neg \textbf{y}$. The negative label comprises of a set of sub-classes e.g. in the context of urban feature classification the negative label \textit{non-building} can correspond to roads, cars, trees, low vegetation, clutter, etc.
    
\end{enumerate}

Given the fact that the deeper convolutional layers contain information about complex objects, we proceed with the experimental proof of the hypothesis by analyzing the penultimate layer before the final logit (softmax) layer which outputs the binary classification of building/non-building.

\subsection{Experimental Proof}
\label{subsec:experimental_proof}

In order to prove the above hypothesis, ground truth for the sub-classes of the negative label is required. In our experiments we use four orthophoto images from the ISPRS benchmark dataset \cite{rottensteiner2012isprs}. For each image $I^{i}, 1 \leq i \leq 4$, a labeled image $I^{i}_{label}$ is also provided as ground-truth showing the manually assigned per-pixel classification into six classes $L$: (a) buildings (blue), (b) roads (white), (c) trees (green), (d) red (clutter), (e) low vegetation/natural ground (cyan), (f) cars (yellow). The 'clutter' class contains areas for which a class could not be assigned e.g. water, vertical walls, etc. The 'low vegetation/natural ground' class contains areas on the ground covered by vegetation other than trees such as low bushes, grass, etc. 

Using the four pairs of images, we first estimate a probabilistic model for each sub-class as shown in Figure \ref{fig:model_estimation}. Next, using these models as probability priors, we make sub-classification predictions at each point by estimating the maximum likelihoods as shown in Figure \ref{fig:sub_classification}. The process is explained in more detail in the sections below.

\subsubsection{Probabilistic Model Estimation}
\label{subsubsec:model_estimation}

We use the pre-trained ICT-Net for inference on the images $I^{i}, 1 \leq i \leq 4$. After each inference, we gather all activation values corresponding to each class $C$ at each feature map of the penultimate layer. To achieve this we use the indices of all the pixels classified as $l \in L$ in the labeled image $I^{i}_{label}$ to gather all the activation values for that class. 
The activation values for each class $l \in L$ are aggregated across the four images $I^{i}$.

\begin{figure}[!ht]
    \centering
    \includegraphics[width=0.5\textwidth]{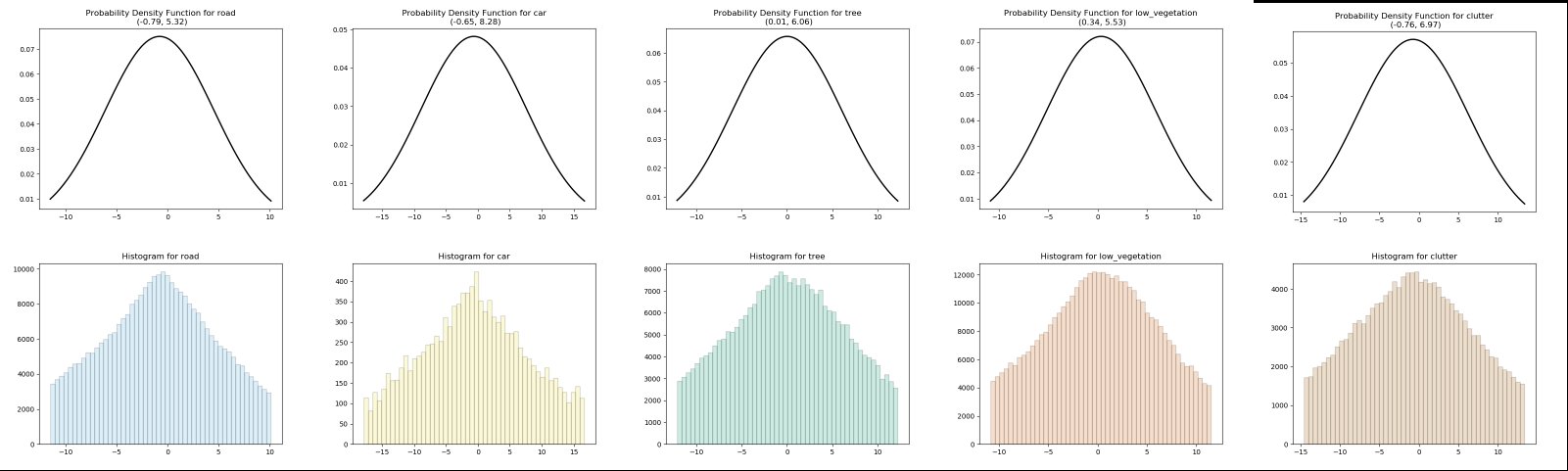}
    \caption{Example histograms of activation values for each sub-class of the negative label. Below each histogram is the PDF of each label modeled with a Gaussian function. Shown are the results for the penultimate layer $\lambda_{n-1}$ before the final logit (softmax) layer $\lambda_{n}$ which outputs the binary classification of building/non-building.}
    \label{fig:histograms}
\end{figure}

An analysis of the histograms of the activation values at each feature map showed that the values resemble a uniform distribution. Thus, a Gaussian function $\mathcal{N}_{C}^{\lambda,\phi}$ is used to model the probability distribution function (PDF) at penultimate layer $\lambda_{n-1}$ and feature map $\phi$. As an example, Figure \ref{fig:histograms} shows the histograms of a randomly selected feature map of the penultimate layer 
$\lambda_{n-1}, \phi=131$, and their corresponding PDFs.

\subsubsection{Sub-classification of the Negative label}
\label{subsubsec:subclassification}
In this step, Maximum-Likelihood Classification (MLC) is used for pixel-wise sub-classification. Thus, every pixel is assigned to the label to which it has the highest probability of being part of. 

The sub-classification is performed at each feature map of the penultimate layer $\lambda_{n-1}$. In the case of ICT-Net this results in a total of $K=256$ classification images $C_{k}, 0 \leq k \leq K$. Each classification image encodes information at the pixel-level about (i) the label that maximizes its probability, and (ii) the probability value. For example, $\langle i, \mathcal{P}(C_{k}^{(x,y)}, \mathcal{N}(\mu_{i},\sigma_{i})) \rangle_{k}$ indicates that the activation value of pixel $(x,y)$ at feature map $C_{k}$ gives the highest probability $\mathcal{P}(C_{k}^{(x,y)}, \mathcal{N}(\mu_{i},\sigma_{i}))$ for label $i$ with PDF $\mathcal{N}(\mu_{i},\sigma_{i})$.


\begin{figure}[!ht]
    \centering
    \includegraphics[width=0.5\textwidth]{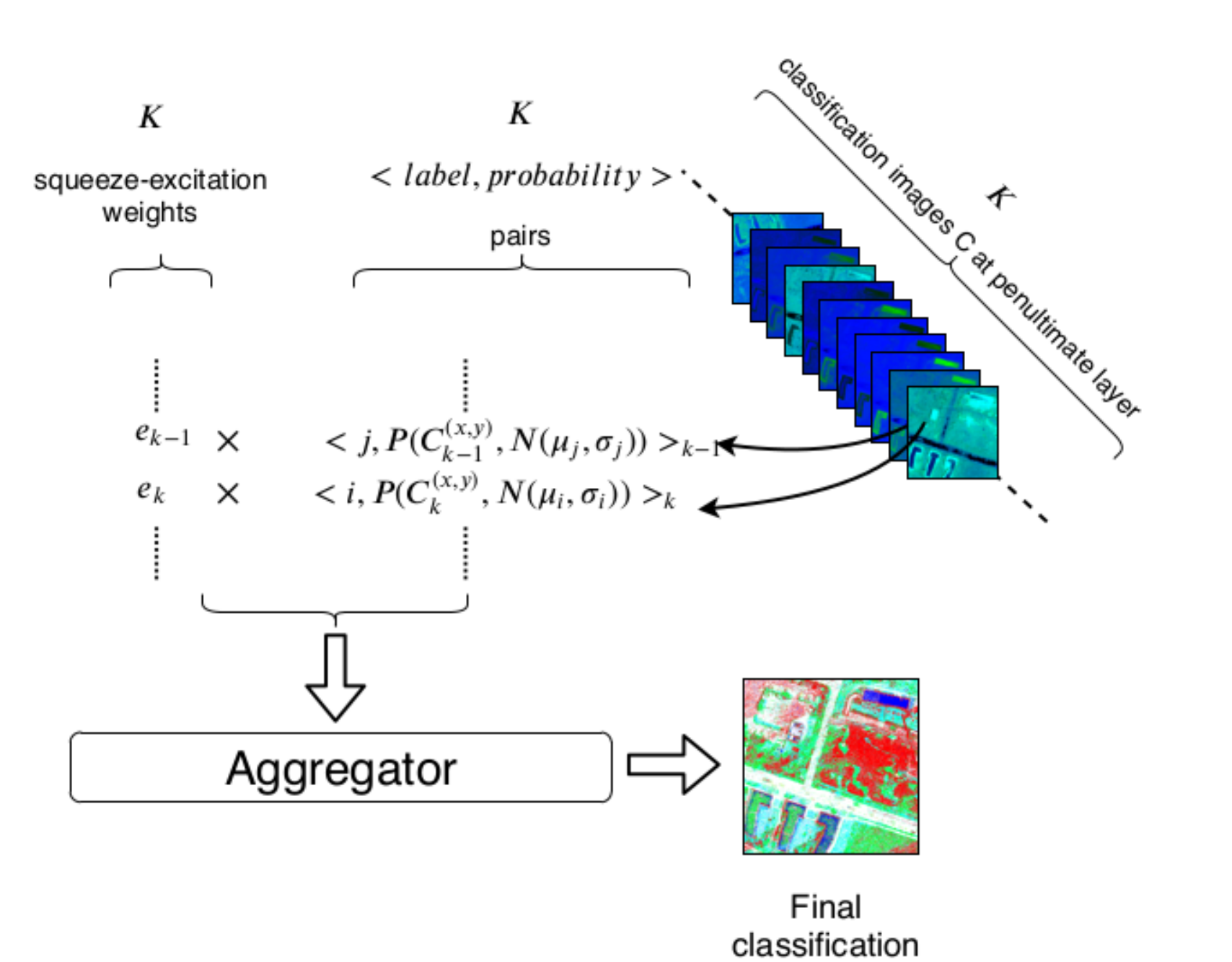}
    \caption{\textbf{Classification aggregator}. There is one classification image per feature map. Each classification image contains pairs of $<label, probability>$ for each pixel. The pairs are converted to one-hot vectors (e.g. vec[$label$] = $probability$) and aggregated together. The label with the highest probability is assigned to each pixel.}
    \label{fig:aggregator}
\end{figure}

\subsubsection{Classification Aggregator}

Next, the $K$ $\langle label, probability \rangle$ pairs are weighted using the squeeze-excitation weights corresponding to each feature map. The weighted pairs for each pixel are converted into one-hot 6-dimensional vectors (e.g. corresponding to the 6 labels) containing only the value of the highest probability $\mathcal{P}(C_{k}^{(x,y)}, \mathcal{N}(\mu_{i},\sigma_{i}))$ at the index of label $i$. Thus, continuing with the previous example, $vec$ is the one-hot 6-dimensional vector corresponding to pixel $(x,y)$ in $C_{k}$ where the only non-zero value is at 
$vec[i] = \mathcal{P}(I_{(x,y)}, \mathcal{N}(\mu_{i},\sigma_{i}))$.

Finally, for each pixel the $K$ one-hot vectors are aggregated and the label for which the total probability is maximal is assigned to the pixel. Figure \ref{fig:aggregator} summarizes the process of the classification aggregator. 

\begin{figure}[!ht]
    \centering
    \begin{subfigure}[b]{0.23\textwidth}
        \includegraphics[width=\textwidth]{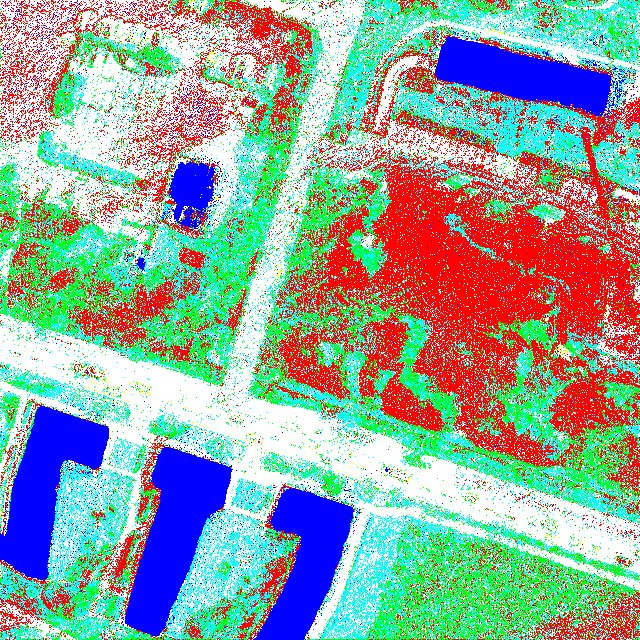}
        \caption{}
        \label{fig:subclassification_mle_output}
    \end{subfigure}
    \hfill
    \begin{subfigure}[b]{0.23\textwidth}
        \includegraphics[width=\textwidth]{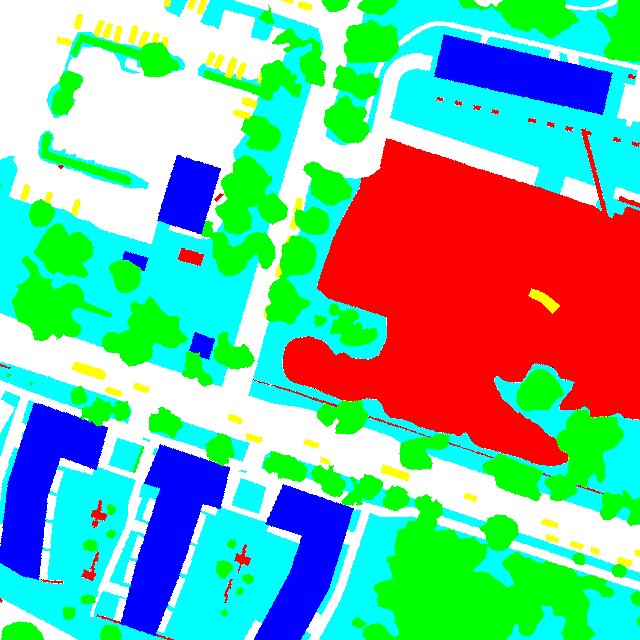}
        \caption{}
        \label{fig:subclassification_mle_gt}
\end{subfigure}
    
    \begin{subfigure}[b]{0.23\textwidth}
        \includegraphics[width=\textwidth]{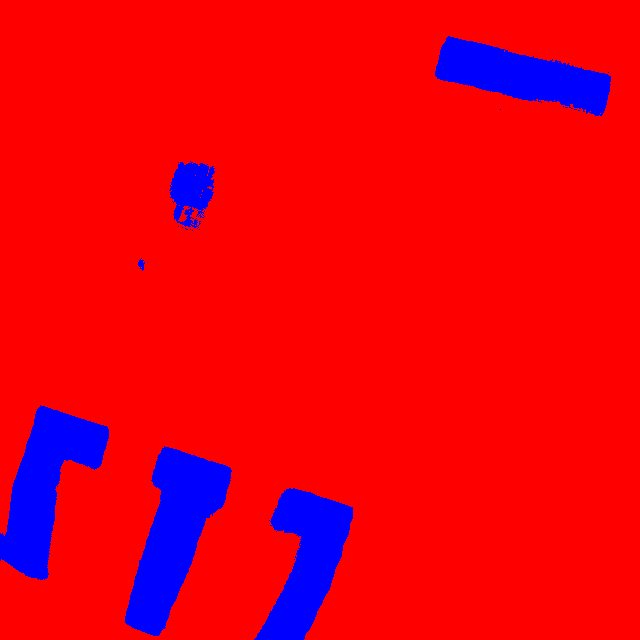}
        \caption{}
        \label{fig:final_classified_features23}
    \end{subfigure}
        \hfill
    \begin{subfigure}[b]{0.23\textwidth}
        \includegraphics[width=\textwidth]{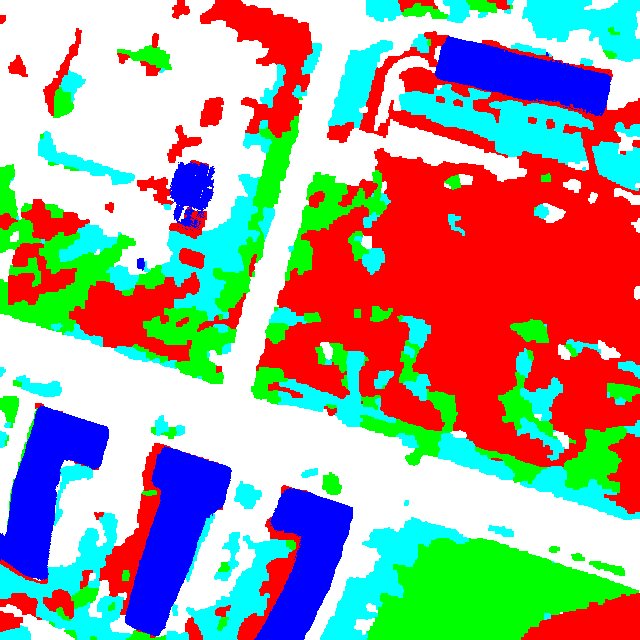}
        \caption{}
        \label{fig:subclassification_mle_output_mrf}
    \end{subfigure}
    \caption{(a) Sub-classification using MLC at penultimate layer (i.e. one before the final logit layer). ICT-Net was not trained nor fine-tuned using any of the images from the ISPRS dataset; yet latent learning is occurring and can be exploited for the sub-classification of the negative label i.e. non-building. (b) Ground-truth. (c) Sub-classification using MLC at the final layer before output. Sub-classes of the negative label are squashed into a single negative label i.e. non-building and separability of their clusters cannot be easily achieved. (d) Refinement of (a) using the technique described in Section \ref{sec:reconstruction_analysis} after updating it to use 6 labels. Since the focus is on sub-classification of the negative label the building predictions from ICT-Net are overlaid on top. 
    }
    \label{fig:sub_classification_mle}
\end{figure}

A final classification result using the activation values of each feature map of the penultimate layer is shown in Figure \ref{fig:subclassification_mle_output}. The ground truth is shown in Figure \ref{fig:subclassification_mle_gt}. Figure \ref{fig:subclassification_mle_output_mrf} shows the result after refinement on \ref{fig:subclassification_mle_output} using the technique described in Section \ref{sec:reconstruction_analysis} after updating it to use 6 labels. Since the focus is on sub-classification of the negative label, we overlaid the binary classification result of ICT-Net on buidings on top of the sub-classification output. 

\begin{table*}[!ht]
    \centering
  \begin{tabular}{|| p{68pt} | p{68pt} | p{68pt} | p{82pt} | p{95pt} | p{52pt} || }
    \hline
    \textbf{Image} & \textbf{\makecell{Multi-label \\ Ground-truth}} & \textbf{\makecell{Building\\prediction}} &
    \textbf{\makecell{Sub-classification of\\ non-building label}} & \textbf{\makecell{Building prediction \\ overlaid on sub-classes}} &
    \textbf{F1 score} (\%) \\
\hline
\includegraphics[width=0.15\textwidth]{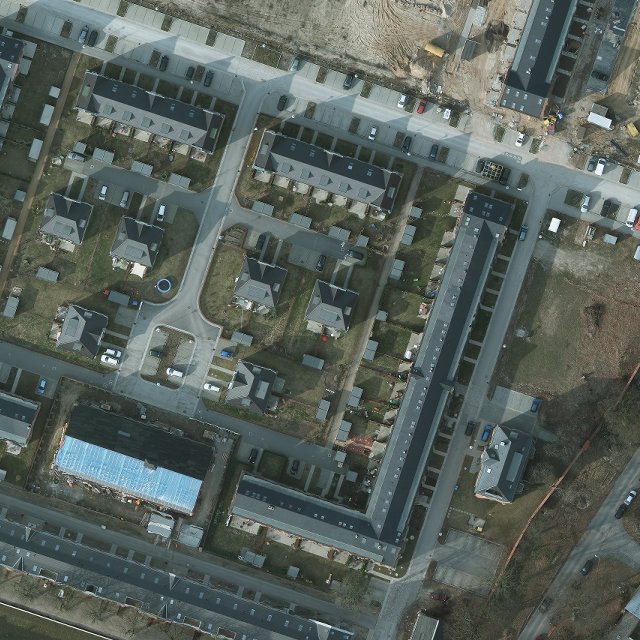}
&
\includegraphics[width=0.15\textwidth]{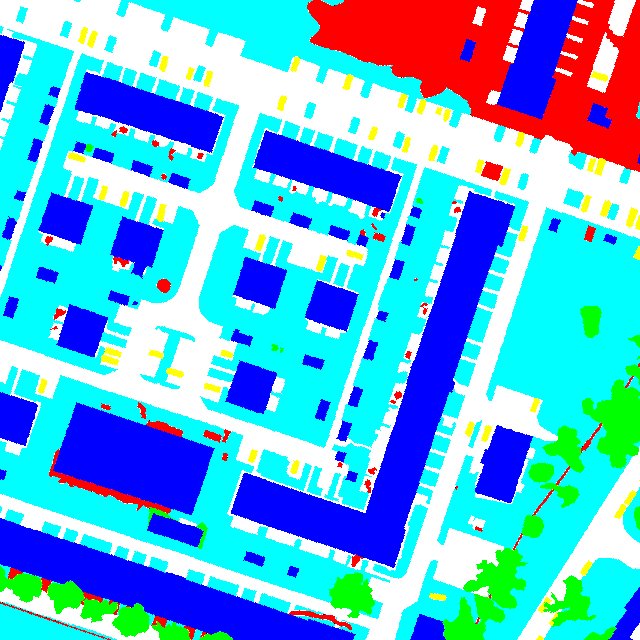}
&
\includegraphics[width=0.15\textwidth]{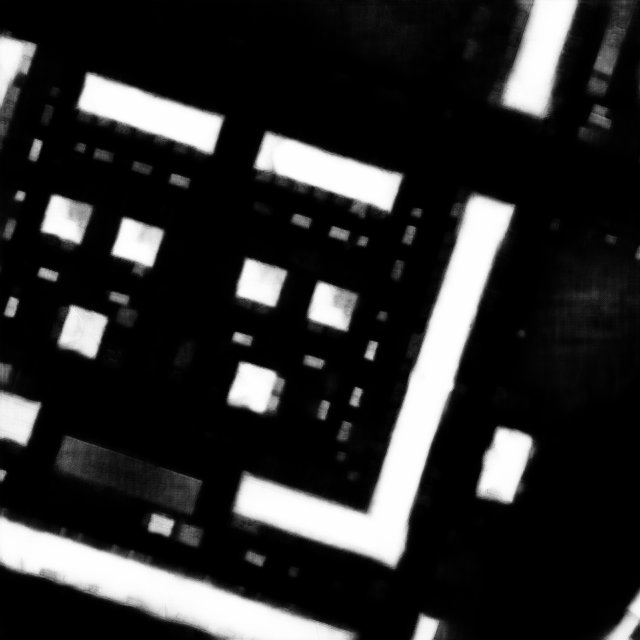}
&
\includegraphics[width=0.15\textwidth]{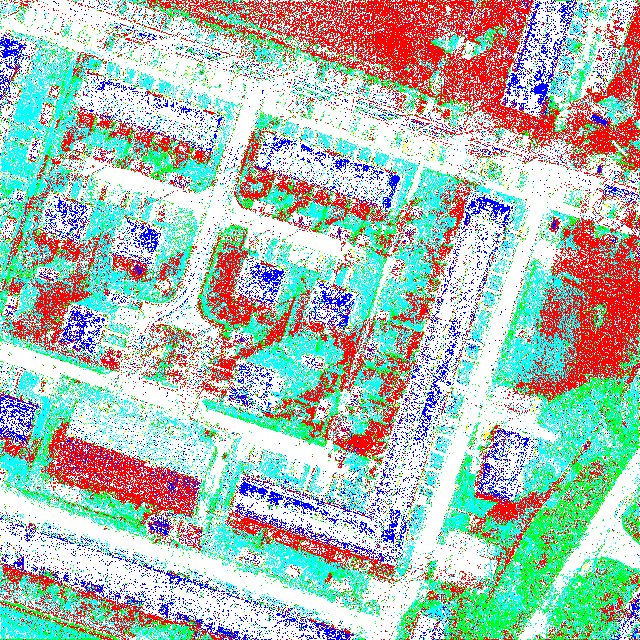}
&
\includegraphics[width=0.15\textwidth]{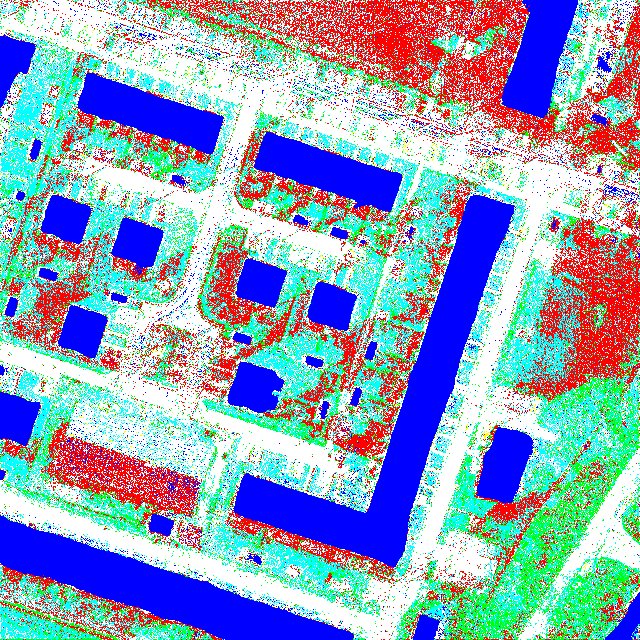}
&
\makecell[b]{B: 0.85\% \\ R: 0.62\% \\ C: 0.08\% \\ T: 0.18\% \\ LV: 0.46\% \\ Cl:0.32\%}
\\
\hline
\includegraphics[width=0.15\textwidth]{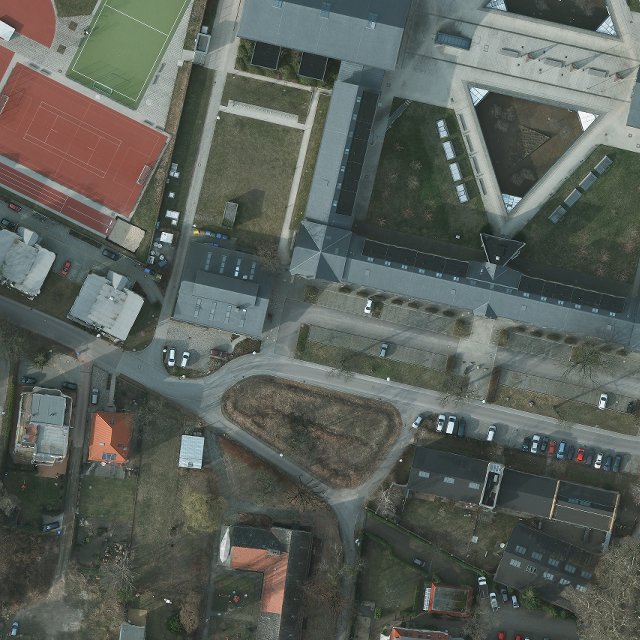}
&
\includegraphics[width=0.15\textwidth]{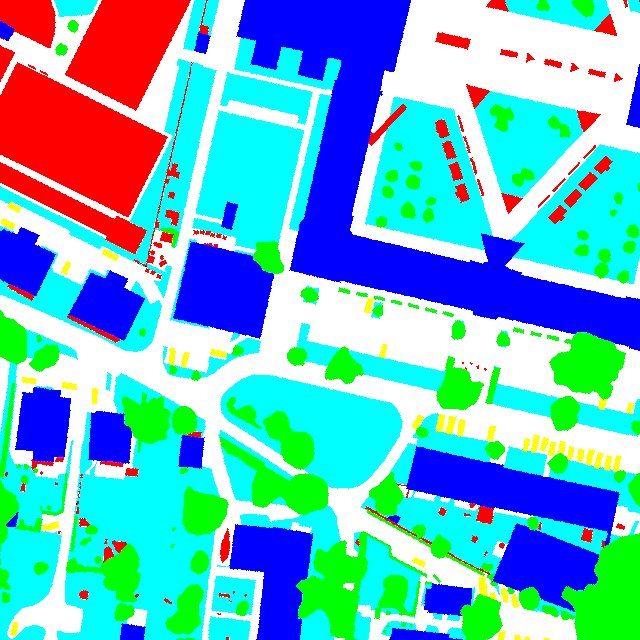}
&
\includegraphics[width=0.15\textwidth]{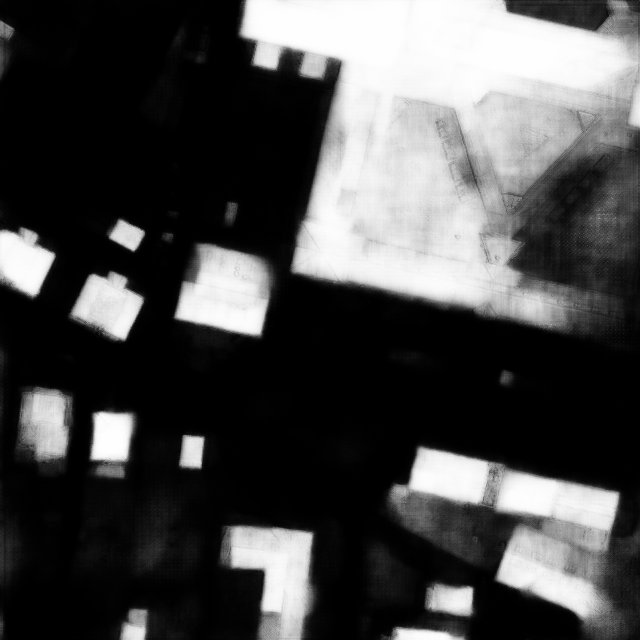}
&
\includegraphics[width=0.15\textwidth]{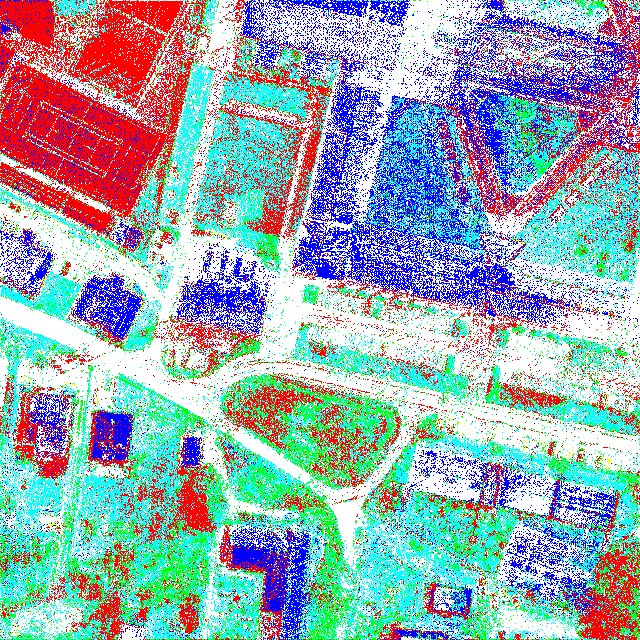}
&
\includegraphics[width=0.15\textwidth]{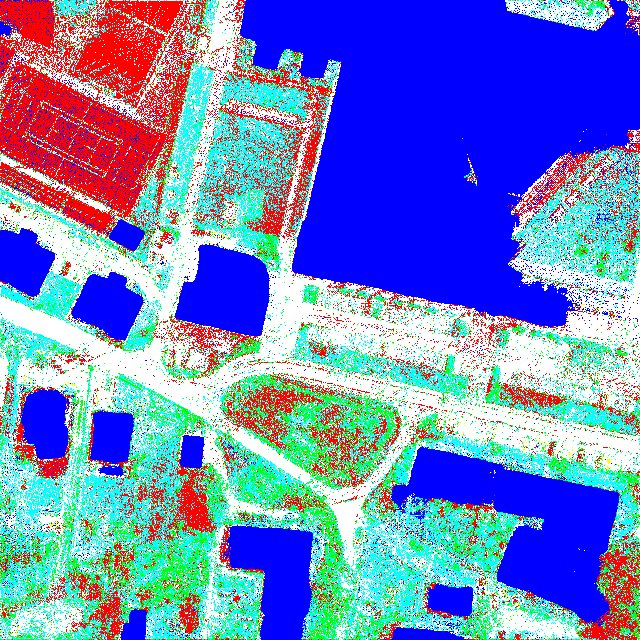}
&
\makecell[b]{B: 0.66\% \\ R: 0.54\% \\ C: 0.09\% \\ T: 0.24\% \\ LV: 0.44\% \\ Cl:0.47\%}
\\
\hline
\includegraphics[width=0.15\textwidth]{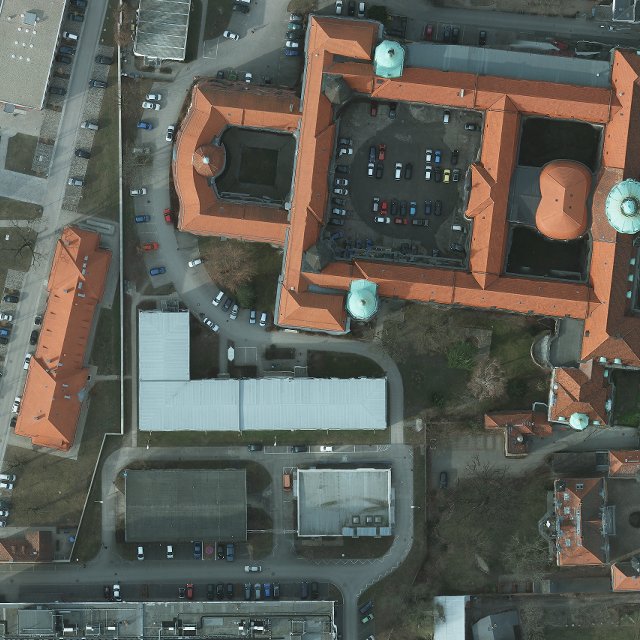}
&
\includegraphics[width=0.15\textwidth]{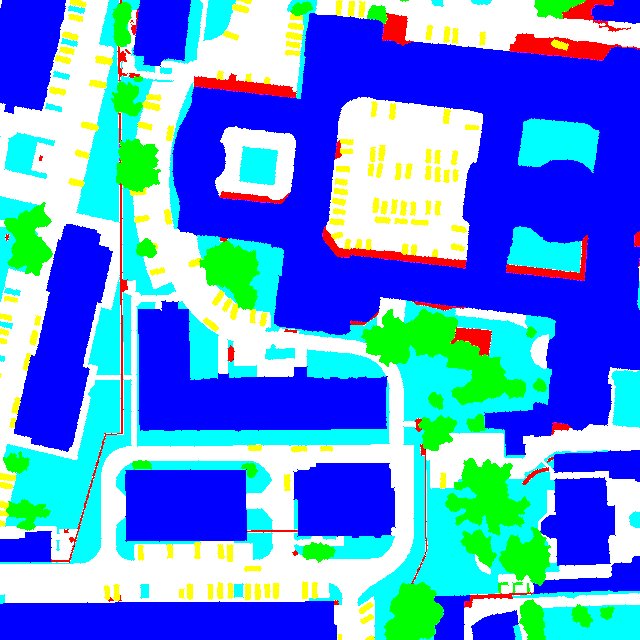}
&
\includegraphics[width=0.15\textwidth]{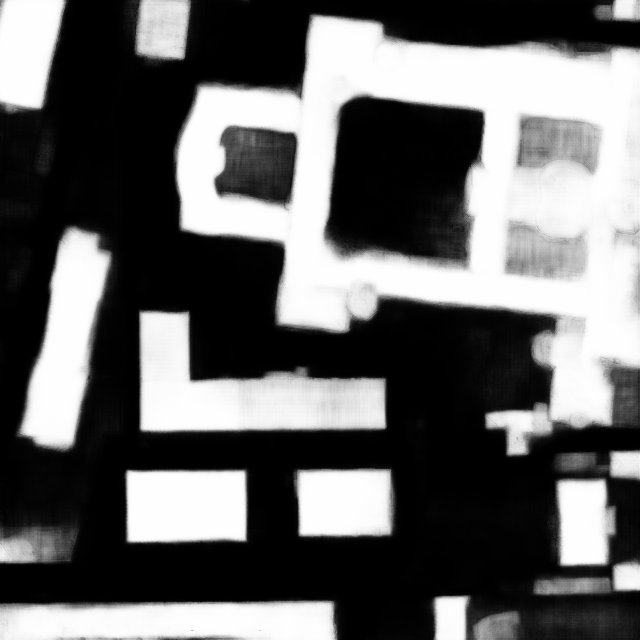}
&
\includegraphics[width=0.15\textwidth]{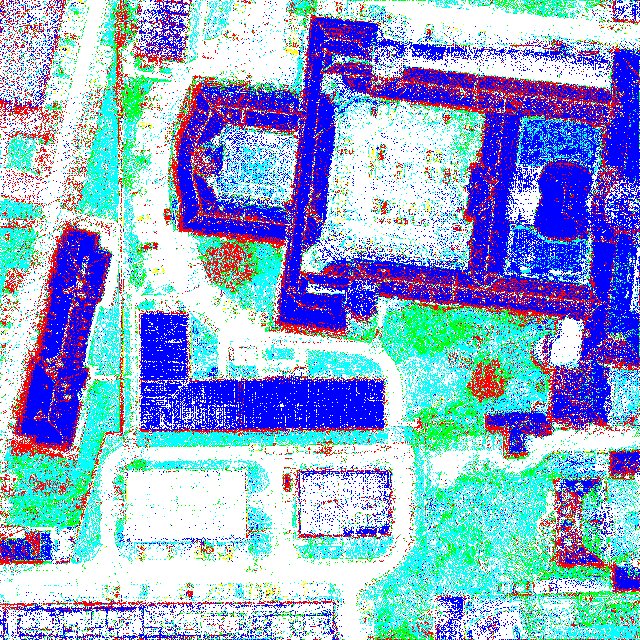}
&
\includegraphics[width=0.15\textwidth]{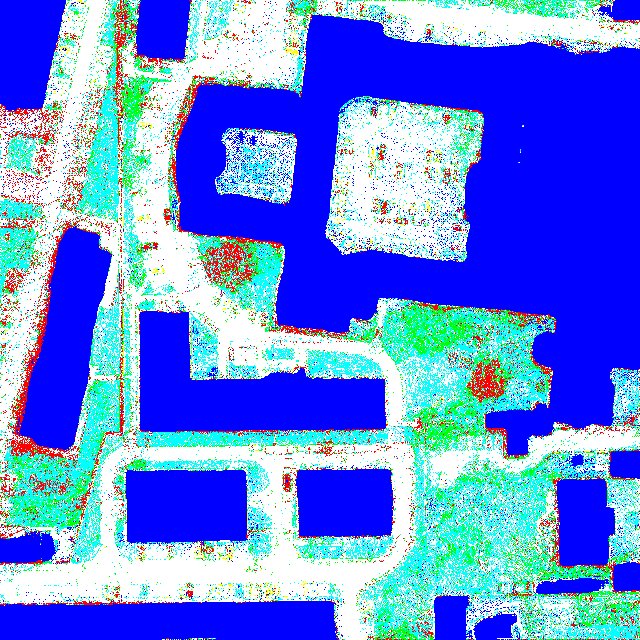}
&
\makecell[b]{B: 0.93\% \\ R: 0.73\% \\ C: 0.12\% \\ T: 0.25\% \\ LV: 0.51\% \\ Cl:0.08\%}
\\
\hline
\includegraphics[width=0.15\textwidth]{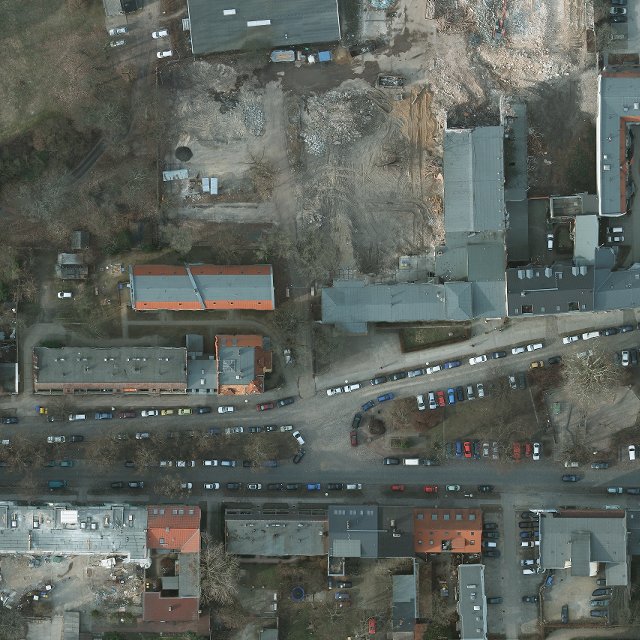}
&
\includegraphics[width=0.15\textwidth]{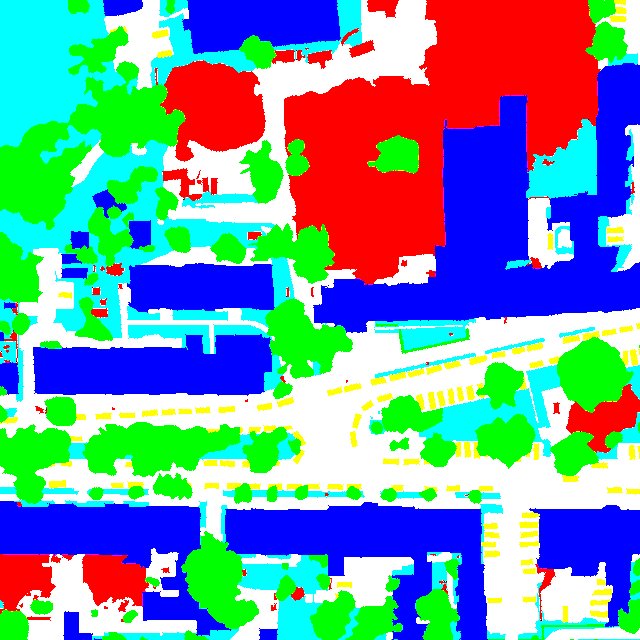}
&
\includegraphics[width=0.15\textwidth]{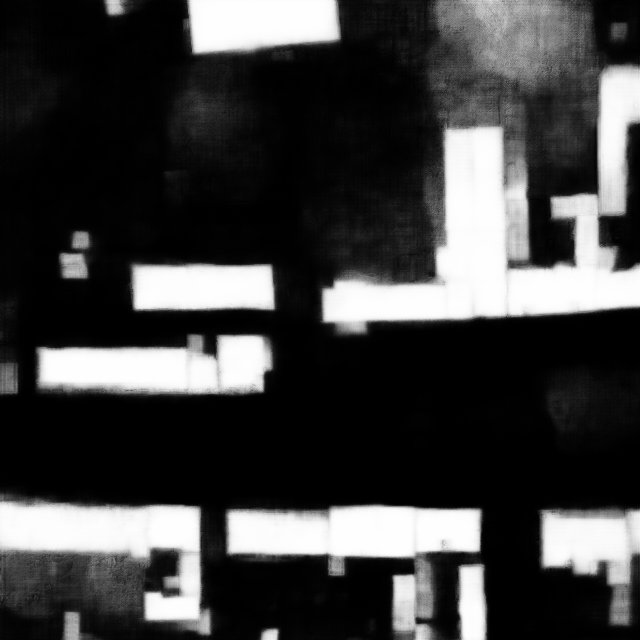}
&
\includegraphics[width=0.15\textwidth]{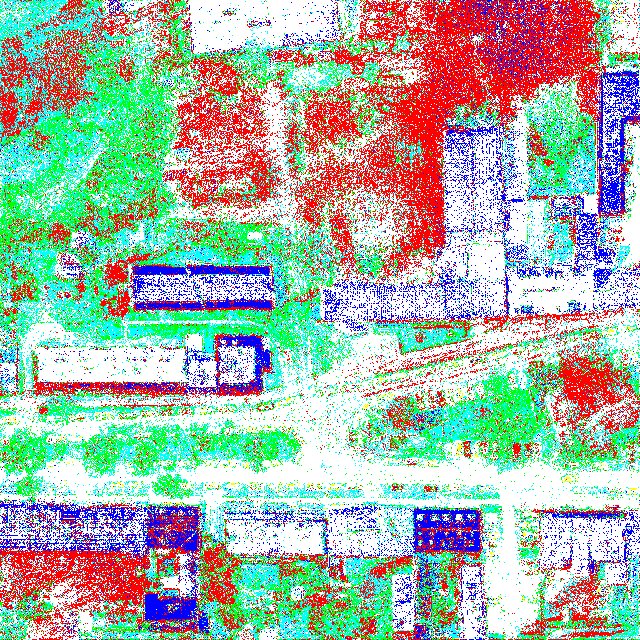}
&
\includegraphics[width=0.15\textwidth]{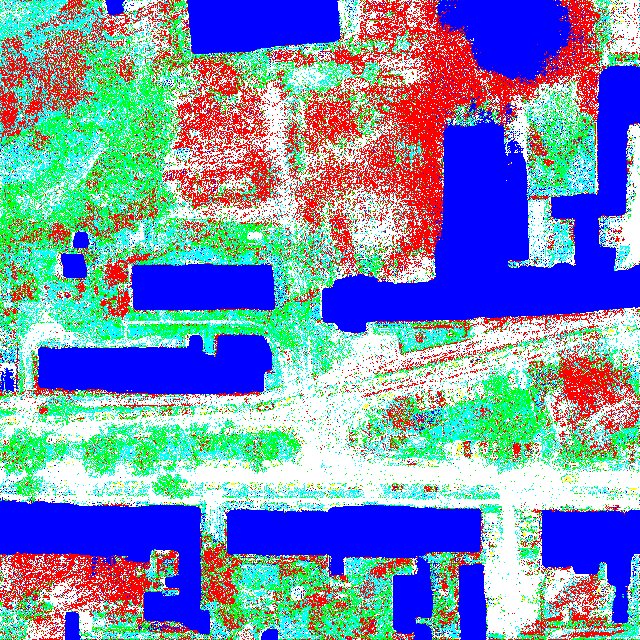}
&
\makecell[b]{B: 0.88\% \\ R: 0.58\% \\ C: 0.12\% \\ T: 0.32\% \\ LV: 0.32\% \\ Cl:0.42\%}
\\
\hline
\includegraphics[width=0.15\textwidth]{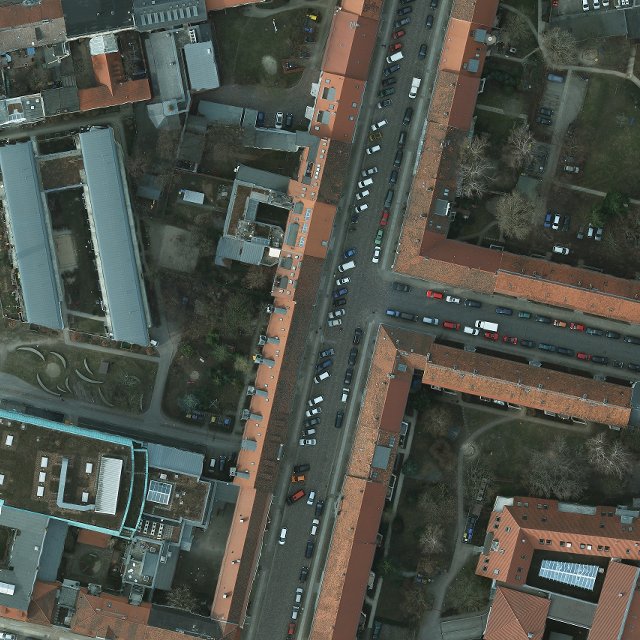}
&
\includegraphics[width=0.15\textwidth]{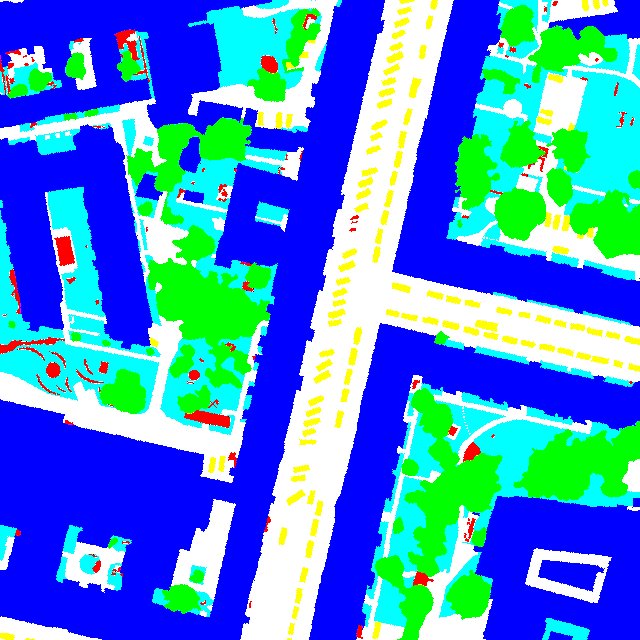}
&
\includegraphics[width=0.15\textwidth]{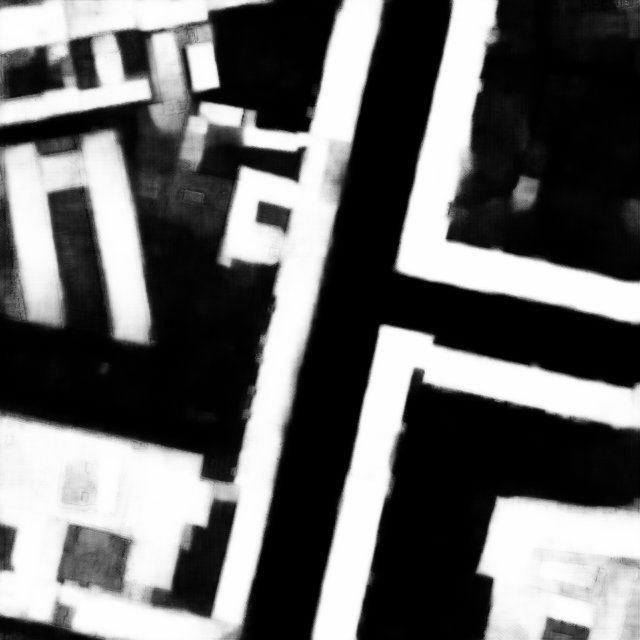}
&
\includegraphics[width=0.15\textwidth]{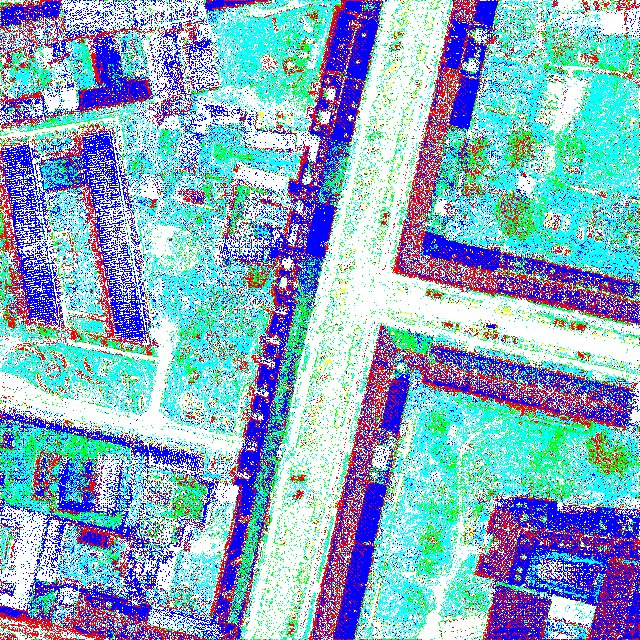}
&
\includegraphics[width=0.15\textwidth]{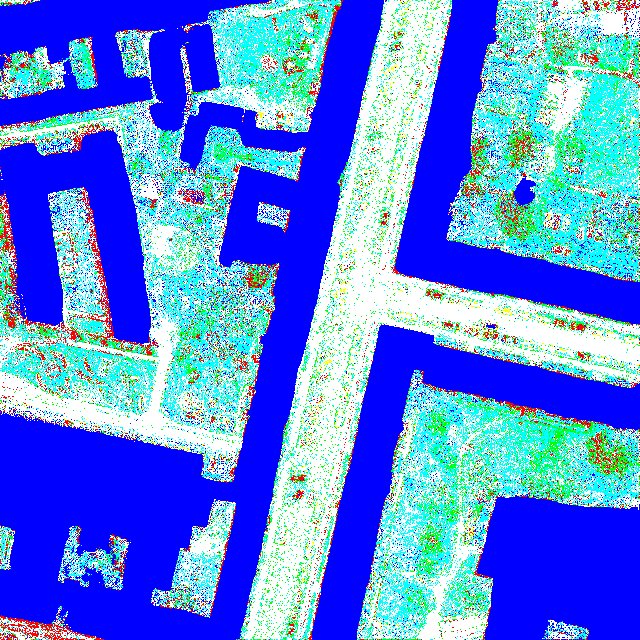}
&
\makecell[b]{B: 0.92\% \\ R: 0.62\% \\ C: 0.13\% \\ T: 0.23\% \\ LV: 0.44\% \\ Cl:0.10\%}
\\
\hline
\includegraphics[width=0.15\textwidth]{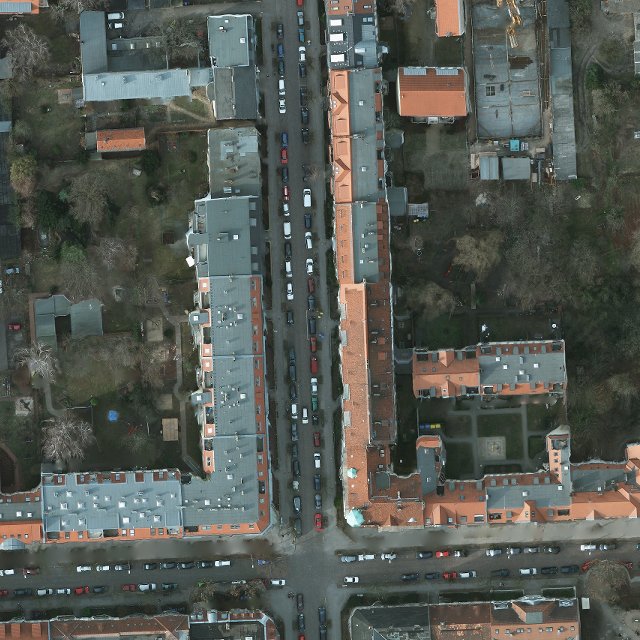}
&
\includegraphics[width=0.15\textwidth]{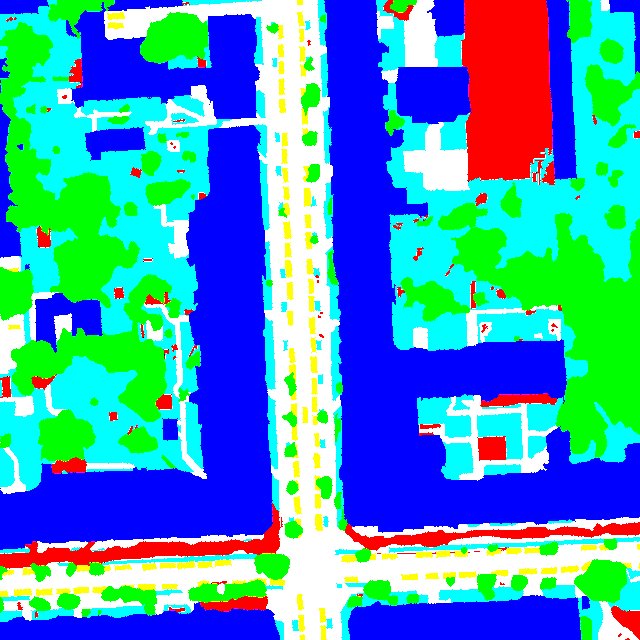}
&
\includegraphics[width=0.15\textwidth]{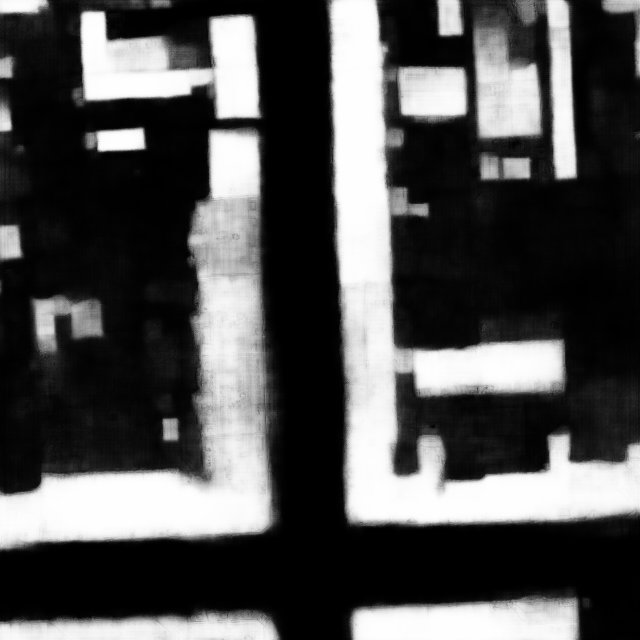}
&
\includegraphics[width=0.15\textwidth]{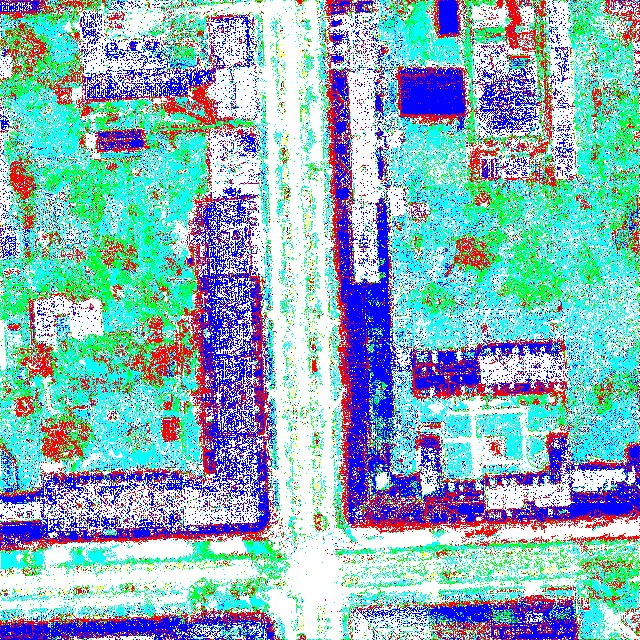}
&
\includegraphics[width=0.15\textwidth]{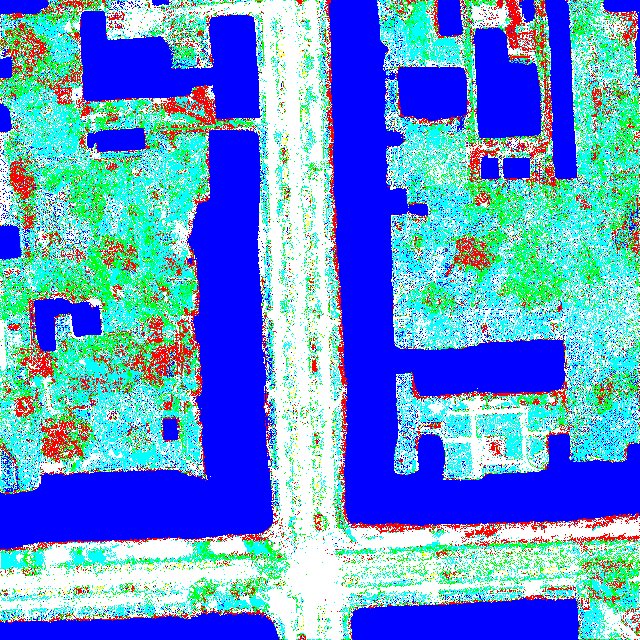}
&
\makecell[b]{B: 0.89\% \\ R: 0.53\% \\ C: 0.11\% \\ T: 0.27\% \\ LV: 0.48\% \\ Cl:0.14\%}
\\
\hline
\includegraphics[width=0.15\textwidth]{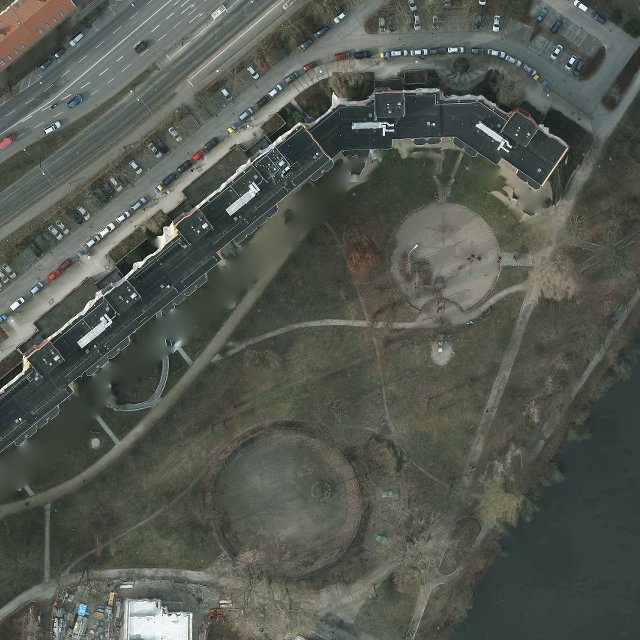}
&
\includegraphics[width=0.15\textwidth]{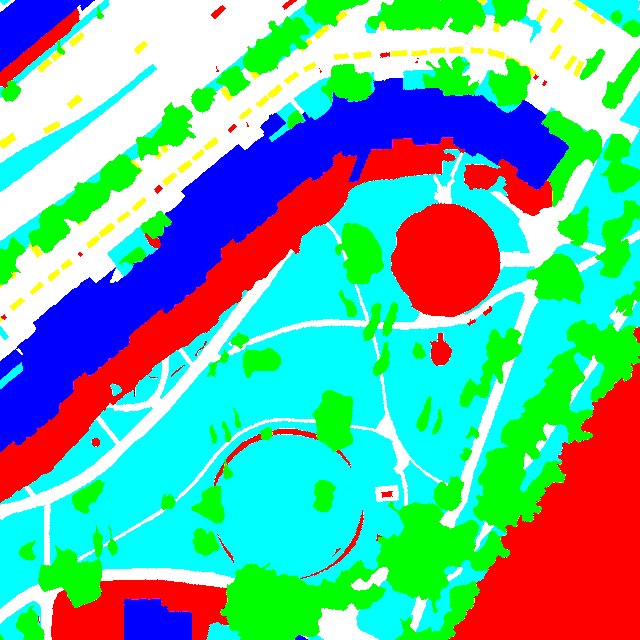}
&
\includegraphics[width=0.15\textwidth]{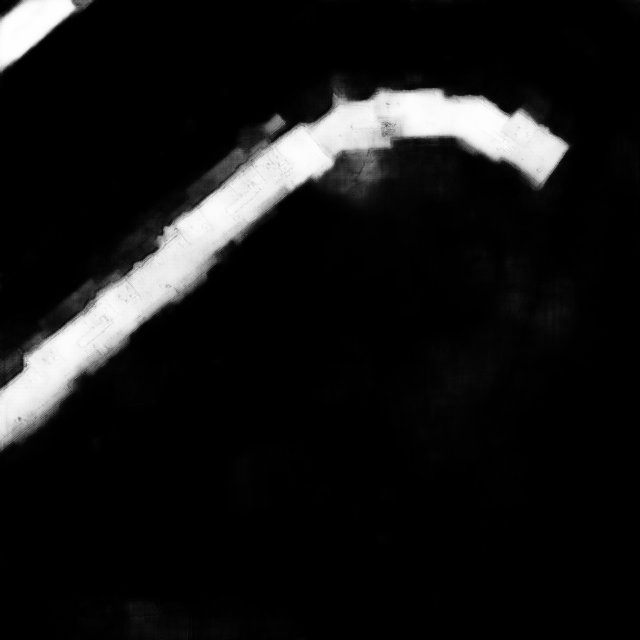}
&
\includegraphics[width=0.15\textwidth]{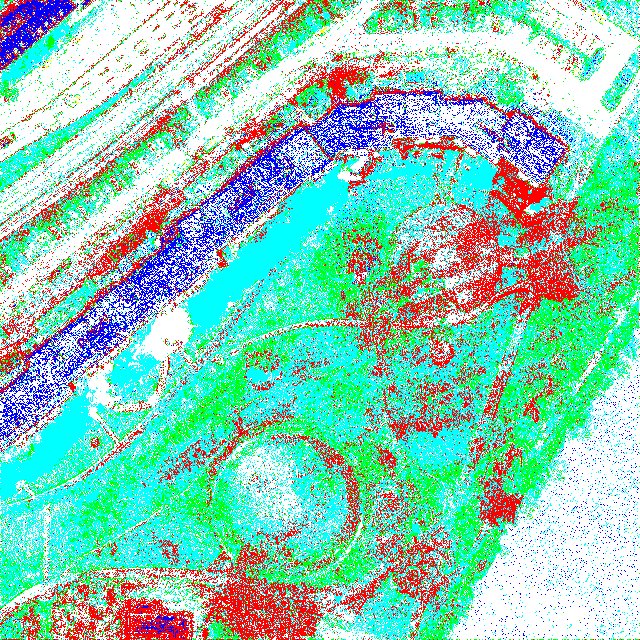}
&
\includegraphics[width=0.15\textwidth]{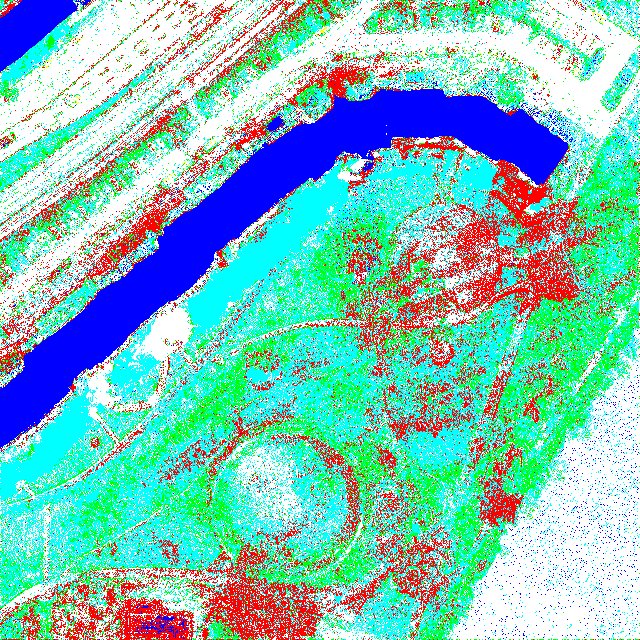}
&
\makecell[b]{B: 0.85\% \\ R: 0.50\% \\ C: 0.08\% \\ T: 0.25\% \\ LV: 0.48\% \\ Cl:0.14\%}
\\
\hline

    \end{tabular}
  \caption{Results on the randomly selected images (first column) from the ISPRS benchmark dataset (Pottsdam). The second column shows the multi-label ground-truth. The third column shows the building predictions from the pretrained ICT-Net. Note that the ISPRS dataset was \textbf{not} used during its training/validation. The fourth column is the sub-classification of the negative label (i.e. non-buildings) using the proposed technique. The fifth column shows the result of overlaying the building predictions (third column) on the sub-classification (fourth column). The sixth column shows the F1 scores for \textbf{B}uilding, \textbf{R}oad, \textbf{C}ar, \textbf{T}ree, \textbf{L}ow \textbf{V}egetation, and \textbf{Cl}utter.}
  \label{tab:results}
\end{table*}

\subsection{Experimental Results}
We have extensively tested the proposed technique on images from the ISPRS benchmark dataset for which multi-label ground truth was available. The results for a number of randomly chosen images are shown in Table \ref{tab:results}. First, the building predictions were generated using the pre-trained ICT-Net (third column). As previously mentioned, the training and validation of the ICT-Net did not include any images from the ISPRS dataset. Next, the sub-classification of the negative label (i.e. non-building) was generated using the proposed technique (fourth column). Since the focus is on sub-classification of the negative label the binary building predictions are overlaid with the blue color on the sub-classifications (fifth column). Table \ref{tab:results} fifth column shows the final composited result which is evaluated using the multi-label ground truth (second column) and the F1-score (sixth column). 

As it can be seen from the results, using the proposed technique one is able to externalize information about the sub-classes of the negative label for which the network did not train on. In particular, for the road (R) and low vegetation (LV) classes the proposed technique achieves much higher performance than one would get by pure chance (i.e. ~16.7\%) which is quantitatively confirmed by the average of the F1 scores on a set of 10 randomly chosen images from the ISPRS benchmark dataset, i.e. R: 54.29\% (SD=17.03), C: 10.15\% (SD=2.54), T: 24.11\% (SD=5.25), LV: 42.74\% (SD=6.62), Cl: 18.30\% (SD=16.08). 

Thus, a network which was trained on a binary classification task using only binary examples of building/non-building, is indeed performing clustering of the sub-classes of the negative label based solely on their similarities and without knowledge on what each sub-class represents. It is worth noting that the F1 scores for the sub-classification of the road (R) and low vegetation (LV) classes are the highest when compared to the scores for the rest of the classes (e.g. cars, trees, clutter). This can be primarily attributed to the fact that roads and low vegetation are integral parts in the building classification process since they are mutually exclusive: a building can never be part of the road nor low vegetation, and vice-versa. This means that misclassifications in these two classes (R, LV) negatively affects the performance on the primary task (building/non-building). As the results clearly show, the network forces the clusters of the activation values corresponding to these two classes to be distinct and separable from the building class. This is further strengthened by the fact that the two classes (R, LV) combined, occupy the majority area within any satellite image compared to the other classes. This observation is also supported by the fact that cars (C), which are very small in size (i.e. number of pixels) and number (i.e. occurrences in an image), have the lowest reported F1 scores.

\subsection{Ablation Study}
\label{subsec:ablation_study}

\begin{table*}[!ht]
    \centering
  \begin{tabular}{|| p{75pt} | p{75pt} | p{25pt} | p{25pt} | p{25pt} | p{25pt} | p{30pt} | p{27pt} | p{45pt} || }
    \hline
    \textbf{Ablations} & \makecell[b]{\textbf{Sub-classification}\\ \textbf{of negative label}\\ \textbf{with building}\\ \textbf{prediction}} & \textbf{B} (\%) &
    \textbf{R} (\%) & \textbf{C} (\%) &
    \textbf{T} (\%) &
    \textbf{LV} (\%) &
    \textbf{Cl} (\%) &
    \textbf{Compute time}\\
    \hline
    \makecell[b]{\textbf{Model:}\\
    GD\\ \textbf{Sub-classification:}\\ MLC} & \includegraphics[width=0.15\textwidth]{images/final_classified_features_22_wo_weights_updated.jpg} & 85.93 & 62.49 & 7.32 & 34.49 & 38.85 & 55.47 & 01h:03m:02s\\ \hline
    \makecell[b]{\textbf{Model:}\\
    GD\\ \textbf{Sub-classification:}\\ MAP-MRF\\ ($w=0.0005$)} & \includegraphics[width=0.15\textwidth]{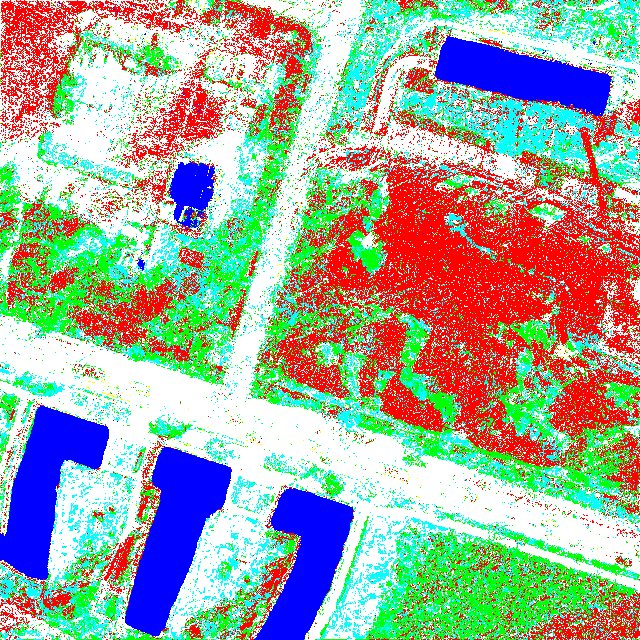} & 90.30 & 59.96 & 5.28 & 35.93 & 31.68 & 54.39 & 01h:32m:47s
    \\ \hline
    \makecell[b]{\textbf{Model:}\\
    GD\\ \textbf{Sub-classification:}\\ MAP-MRF\\ ($w=0.005$)} & \includegraphics[width=0.15\textwidth]{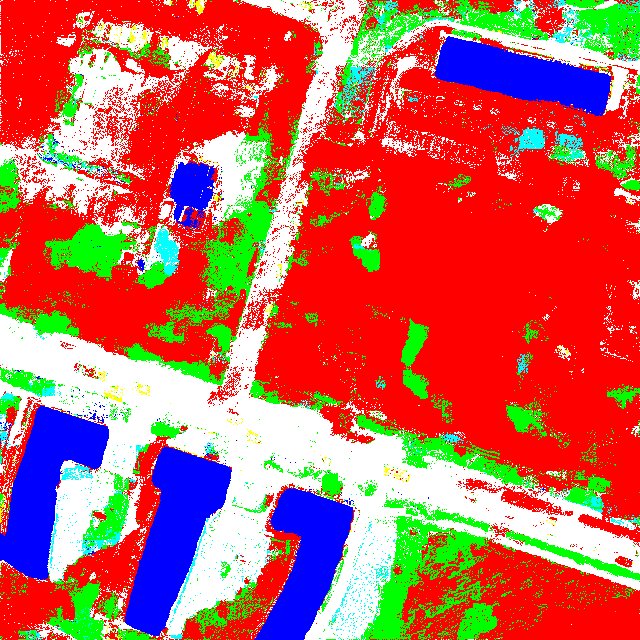} & 89.97 & 56.47 & 18.42 & 35.35 & 7.79 & 48.17 & 01h:28m:19s \\ \hline
    \makecell[b]{\textbf{Model:}\\
    GD\\ \textbf{Sub-classification:}\\ MAP-MRF\\ ($w=0.05$)} & \includegraphics[width=0.15\textwidth]{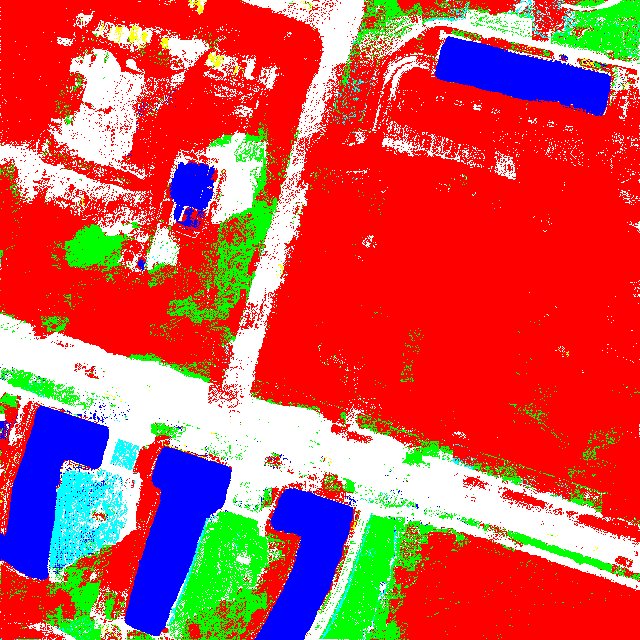} & 89.44 & 60.42 & 15.17 & 20.12 & 9.02 & 46.87 & 01h:24m:31s\\ \hline
    \makecell[b]{\textbf{Model:}\\
    GD\\ \textbf{Sub-classification:}\\ MAP-MRF\\ ($w=0.1$)} & \includegraphics[width=0.15\textwidth]{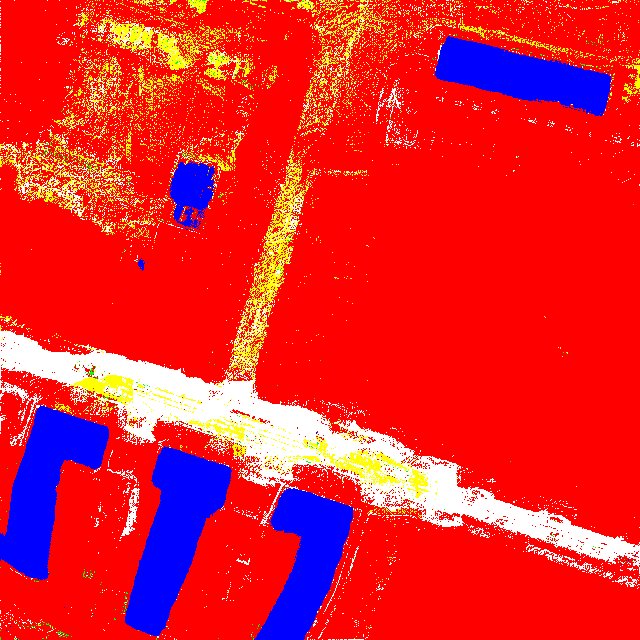} & 90.62 & 37.74 & 76.98 & 0.39 & 0.0 & 37.99 & 01h:35m:55s\\ \hline
\end{tabular}
  \caption{\textbf{Ablations.} F1 scores for \textbf{B}uilding, \textbf{R}oad, \textbf{C}ar, \textbf{T}ree, \textbf{L}ow \textbf{V}egetation, and \textbf{Cl}utter. The bottom four rows show the results of using an alternative method for sub-classification i.e. MAP-MRF, each with a  different weight $w$ in equation \ref{eq:both_terms}.}
  \label{tab:ablations}
\end{table*}



As part of the ablation study we have performed the following experiments. Instead of using MLC, we explored the use of MAP estimation for MRF as a classifier. 
    
    \textbf{Procedure:} The estimated probabilistic models of each class are used as priors in a maximum-a-posterior (MAP) estimation. MAP inference is performed using Graph-cut energy minimization for optimizing the cost function given by,
\begin{equation}
E(f) = E_{unary}(f) + w \times E_{pairwise}(f)
\label{eq:both_terms}
\end{equation}
where $f : I_{p} \rightarrow L$ is an optimal labeling that assigns a label $l \in L$ to each pixel $p \in I$, and $w$ is the weight indicating the importance of one term with respect to the other.

The unary term $E_{unary}$ provides a per-pixel estimate of how appropriate a label $l \in L$ is, for a pixel $p \in I$ in the observed data. The probabilistic models of each class at each layer/feature map are used as priors for calculating this term and is given by,
\begin{equation}
	E_{unary}(f) = \sum_{p \in I} \left[\frac{1}{1 + \mathcal{N}^{l}_{(\lambda,\phi)} (\upsilon_{(\lambda,\phi)}^{p}| \mu_{(\lambda,\phi)}, \sigma_{(\lambda,\phi)})}\right]
\end{equation}
where $\upsilon_{(\lambda,\phi)}^{p}$ is the activation value of pixel $p$ at layer $\lambda$ and feature map $\phi$,  $\mathcal{N}^{l}_{(\lambda,\phi)} ( \mu_{(\lambda,\phi)}, \sigma_{(\lambda,\phi)})$ is the probability prior for label $l$.

The pairwise term $E_{pairwise}$ provides an estimate of the similarity of the labels of two neighbouring pixels $p, q \in I$ with labels $l_{p} , l_{q} \in L$ respectively. Let $f(p)$ and $f(q)$ be the new labeling under $f$ for pixels $p, q \in I$, respectively. $E_{pairwise}(f)$ at layer $\lambda$ and feature map $\phi$ is then defined as, 

\begin{equation}
E_{pairwise}(f) = \sum_{ \{p,q\} \in N}
    \begin{cases}
        40,& \text{if } l_{p} \neq l_{q}\\
        \Delta(f_{p}, f_{q}), & \text{otherwise}
    \end{cases}
\end{equation}
where $N$ is the local 4-neighbourhood, and $\Delta(f_{p}, f_{q})$ is defined as the difference of the probability priors of $f_{p}$, $f_{q}$ given by, 
\begin{align}
\hspace{-10pt}
\begin{split}
  \Delta(f_{p}, f_{q}) =  \mathcal{N}^{f(p)}_{(\lambda,\phi)} (\upsilon_{(\lambda,\phi)}^{p}| \mu_{(\lambda,\phi)}, \sigma_{(\lambda,\phi)}) - \\ \mathcal{N}^{f(q)}_{(\lambda,\phi)} (\upsilon_{(\lambda,\phi)}^{q}| \mu_{(\lambda,\phi)}, \sigma_{(\lambda,\phi)})       
\end{split}
\end{align}

Table \ref{tab:ablations} shows quantivative and qualitative comparisons of the results between MLC and MAP-MRF for different weights $w$ from equation \ref{eq:both_terms}. With respect to the sub-classes of the negative labels (e.g. roads, cars, trees, low vegetation, and clutter) the MLC outperforms MAP-MRF F1 scores in all sub-classes except cars. At the same time, the computational time increases by an average of 00h:30m:38s (experiments were ran on a quad-core Intel Core i7 64-bit machine with 16GB of RAM). 




\subsection{Discussion}

\textbf{Internal validity.} The fact that the activation values of each label are separable and that using a single Gaussian function can accurately model the PDF of every label at each layer $\lambda$ and feature map $\phi$ is a clear indication that the network is performing multi-label clustering. In the case of binary classification such as ours, we showed that this clustering involves labels corresponding to objects of similar characteristics (e.g. shape, color, etc) which even at the penultimate layer are distinct and separable. At the last layer this multi-label information is squashed into a single negative label i.e. non-building. This can be confirmed by comparing the sub-classification results using the PDFs at the penultimate layer shown in Figure \ref{fig:subclassification_mle_output}, and the final logit layer shown in Figure \ref{fig:final_classified_features23}. As it can be seen, the pre-activated convolutions of the final layer have transformed the activation values of the sub-classes of the negative label such that separability of their clusters cannot be easily achieved; leading to the majority of the non-building pixels to be classified with the same label e.g. clutter (red).

\noindent
\textbf{Outliers.} The pre-activated convolutions of ICT-Net include as part of it the activation function ReLU which maps all negative activation values to zero. This can lead to a large number of zero values at each layer/feature map which should not be taken into account when estimating the probabilistic model. Thus, outlier removal is applied at each layer/feature map to remove these zero-values from all gathered activation values per-label prior to fitting a Gaussian distribution.

\noindent
\textbf{Ground Sampling Density (GSD).} ICT-Net is trained on orthophotos with a GSD of $30cm$. Our experiments in Table \ref{tab:variation_of_gsd} show that some variations in GSD can be tolerated by ICT-Net without greatly impacting the performance. However, in the case of the ISPRS benchmark dataset the GSD is $5cm$ which is significantly smaller. Therefore, for the experiments on latent learning we first convert the four orthophotos and ground truth images to GSD $30cm$ by down-sampling.

\begin{table}[!ht]
    \centering
  \begin{tabular}{|| l | c | c || }
    \hline
    \textbf{Ground Sampling Density} & \textbf{IoU} (\%) & \textbf{Accuracy} (\%) \\ \hline
    5 cm & 43.64 & 84.44 \\ \hline
    10 cm & 72.68 & 92.25 \\ \hline
    20 cm & 81.13 & 94.55 \\ \hline
    30 cm & 78.19 & 93.84 \\ \hline
    40 cm & 81.14 & 94.69 \\ \hline
    50 cm & 78.92 & 94.04 \\ \hline
    60 cm & 77.03 & 93.47 \\ \hline
  \end{tabular}
  \caption{Performance evaluation of ICT-Net (pretrained on 30cm INRIA dataset) on multiple ground sampling density of ISPRS potsdam dataset created by down-sampling the orginal orthophoto RGB imagery. Original ground sampling density of ISPRS potsdam dataset was 5cm. The evaluation is performed after converting the multi-labels ground truth into building/non-building per-pixel. IoU refers to IoU of the building class, and Accuracy is the per-pixel accuracy.}
  \label{tab:variation_of_gsd}
\end{table}

\noindent
\textbf{SE weights.} - While analyzing the activations at different layers of the network we extracted the activations of dense blocks and the corresponding excitation values produced by the Squeeze and Excitation block. It is interesting to observe that the excitation values produced by the network for all but last dense block's activations were binary in nature, which leads to a lot of feature maps having zero values and not contributing to the subsequent inference. It is for this reason that the excitation values are used as importance weights to modulate the $<label, probability>$ pairs by the classification aggregator as shown in Figure \ref{fig:aggregator}. 

\noindent
\textbf{Zero-shot learning.} The proposed technique is related to - but different from - zero-shot learning. In contrast to zero-shot learning where the objective is to classify \textit{unseen} classes based on embedding-vectors, our objective is to classify \textit{seen} classes which are unlabeled but appear in the training examples and for which no further information is available. 

\noindent
\textbf{Few-shot learning.} Our technique has more similarities with the objective of few-shot learning. In few-shot learning a small number of images is available and labeled. Meta-learning is used for adapting a pre-trained network to a new task. 
The main difference with our proposed technique is that in our case there is no additional training occurring. A minimal labeled dataset is used for estimating the probabilistic models. The pre-trained network remains unchanged and is used only for inference without additional training for adaptation to the new tasks. Furthermore, the "seen" unlabeled classes coappear in the images with the labeled class.


\vspace{-10pt}
\section{Conclusion}
\label{sec:conclusion}
We have addressed three aspects of semantic segmentation from remote sensor data. We have presented a novel network which combines the strengths of state of the art techniques like Dense blocks in fully convolutional networks and feature recalibration using SE blocks. We have identified the requirements for the particular task and based our decisions on the actual characteristics and observations. We have shown that the proposed architecture outperforms other state of the art including ensemble techniques. 

Furthermore, we investigated the relation between the classification accuracy and the reconstruction accuracy. Due to the extreme difficulty of acquiring blueprints for such large areas and the unavailability of 3D information we have used the building boundaries as a proxy to the reconstruction accuracy. The proposed ICT-Net at different training snapshots was used to generate binary maps of different classification accuracies which were then used for extracting the boundaries.  We presented a comparative quantitative analysis which shows a strong correlation between the two but also a consistent and considerable decrease of the reconstruction accuracy when compared to the classification accuracy. 

Finally, we introduce the concept of latent learning in the context of deep neural network. We experimentally showed that a pre-trained network on binary classification task unintentionally learns about auxiliary tasks and presented a technique for externalizing this knowledge by performing sub-classification of the negative label. Using the proposed technique we demonstrated that sub-classification is possible with acceptable accuracies. This technique is ideal for cases where labels are not easily obtainable but labels for a complementary task are available. 


With respect to the future work, we plan on extending the work (i) by designing a loss function that incorporates the reconstruction accuracy in addition to the classification accuracy during training, and (ii) by applying the concept of latent learning on deep neural networks of different architectures.


\vspace{-10pt}
\ifCLASSOPTIONcompsoc
  \section*{Acknowledgments}
\else
  \section*{Acknowledgment}
  This research is based upon work supported by the Natural Sciences and Engineering Research Council of Canada Grants DG-N01670 (Discovery Grant) and DND-N01885 (Collaborative Research and Development with the Department of National Defence Grant). The authors would like to thank Jonathan Fournier from Valcartier DRDC, and Hermann Brassard, Sylvain Pronovost, Dave Lajoie, and Xian Wang from Presagis Inc Canada, for their invaluable discussions and assistance in processing maps for Montreal. The authors would also like to thank Defence Research and Development Canada and Thales Canada for providing orthophoto RGB images used for testing, and the reviewers for their comments and suggestions.

\fi

This research is supported in part by the Natural Sciences and Engineering Research Council of Canada Grants DG-N01670 (Discovery Grant) and DND-N01885 (Collaborative Research and Development with the Department of National Defence Grant).

\ifCLASSOPTIONcaptionsoff
  \newpage
\fi




\bibliographystyle{abbrv}
\bibliography{main}
%

%

%

\begin{IEEEbiography}[{\includegraphics[width=1in,height=1in,clip,keepaspectratio]{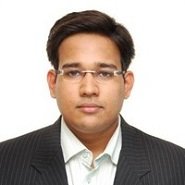}}]{Bodhiswatta Chatterjee} was born in Calcutta, India, in 1987. He received the B.Tech degree in Computer Science and Engineering from Kolkata, India, in 2011, and joined Tata Consultancy Services (TCS) India as a Software Engineer in the same year. In his industrial career of 6 years at TCS, he worked on multiple technologies, and was promoted to a senior Software Engineer in 2015. In 2017, he moved to Montreal, Canada to pursue a Master degree at Concordia University. He completed his M.Sc. in Computer Science with specialisation in Computer Vision from Concordia University, Montreal, Canada, in 2019. He is currently working as a computer vision researcher at Presagis Inc Canada.

His current research interests lie at the intersection of Computer Vision and Deep Neural Networks. More specifically, he is involved in fundamental and applied research covering the following areas: Feature Extraction \& Classification, 3D reconstruction, and Interpretablity in Deep Neural Networks. Bodhiswatta is a member of the Canadian Artificial Intelligence Association (CAIAC) and a student member of the Institute of Electrical and Electronics Engineers (IEEE) Computer Society.
\end{IEEEbiography}

\begin{IEEEbiography}[{\includegraphics[width=1in,height=1in,clip,keepaspectratio]{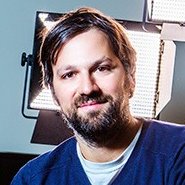}}]{Charalambos Poullis} was born in Nicosia, Cyprus, in 1978. He received the B.Sc. degree in Computing and Information Systems with First Class Honors from the University of Manchester, UK, in 2001, and the M.Sc. in Computing Science with specialization in Multimedia and Creative Technologies, and Ph.D. in Computer Science from the University of Southern California (USC), Los Angeles, USA, in 2003 and 2008, respectively. He is an Associate Professor with the Department of Computer Science and Software Engineering at the Gina Cody School of Engineering and Computer Science at Concordia University where he also serves as the Director of the Immersive and Creative Technologies (ICT) lab. 

His current research interests lie at the intersection of computer vision and computer graphics. More specifically, he is involved in fundamental and applied research covering the following areas: feature extraction \& classification, acquition technologies \& 3D reconstruction, photo-realistic rendering, virtual \& augmented reality. 

Charalambos is a senior member of the Institute of Electrical and Electronics Engineers (IEEE) Computer Society and a member of the Association for Computing Machinery(ACM); Marie Curie Alumni Association (MCAA); British Machine Vision Association (BMVA). He has been serving as a regular reviewer in numerous premier conferences and journals since 2003. 
\end{IEEEbiography}

\vfill




\end{document}